\pdfoutput=1
\documentclass[11pt]{article}
\PassOptionsToPackage{table}{xcolor}

\usepackage[final]{acl}
\usepackage{times}
\usepackage{latexsym}
\usepackage{epigraph}
\usepackage[T1]{fontenc}
\usepackage[utf8]{inputenc}
\usepackage{microtype}
\usepackage{inconsolata}

\usepackage{graphicx}

\usepackage{xspace}
\usepackage{multirow}
\usepackage{booktabs}
\usepackage{bm}
\usepackage{amsmath}
\usepackage{amsfonts}
\usepackage{hyperref}
\usepackage{graphicx}
\usepackage{framed}
\usepackage{enumitem}
\usepackage{pifont}
\usepackage{CJKutf8}
\usepackage{marvosym}
\usepackage{lipsum}
\usepackage{CJKutf8}
\usepackage{pmboxdraw}
\usepackage{caption}
\usepackage{subcaption}
\usepackage{epigraph}
\usepackage{longtable}

\newcommand{\ignore}[1]{}

\usepackage[normalem]{ulem}
\useunder{\uline}{\ul}{}
\definecolor{background_u}{RGB}{240, 240, 240}
\definecolor{frame_u}{RGB}{200, 200, 200}
\definecolor{define_pink}{rgb}{0.988, 0.541, 0.416} 
\definecolor{define_gray}{rgb}{0.502, 0.502, 0.502}

\definecolor{background_u}{HTML}{FEF9F5}
\definecolor{frame_u}{HTML}{98450E}
\definecolor{background_i}{HTML}{F9FBFD}
\definecolor{frame_i}{HTML}{2E75B5}
\definecolor{background_e}{HTML}{FAFAFA}
\definecolor{frame_e}{HTML}{0D0D0D}

\makeatletter

\def\@fnsymbol#1{}

\makeatother

\title{From Neurons to Semantics: Evaluating Cross-Linguistic Alignment \\Capabilities of Large Language Models via Neurons Alignment}

\author{
Chongxuan Huang\textsuperscript{\rm{1,3}}, 
Yongshi Ye\textsuperscript{\rm{2,3}}, 
Biao Fu\textsuperscript{\rm{2,3 \Letter}}, 
Qifeng Su\textsuperscript{\rm{1,3}},
Xiaodong Shi\textsuperscript{\rm{1,3 \Letter}}\thanks{\textsuperscript{\Letter}\ Corresponding author} \\
\textsuperscript{1} School of Informatics, Xiamen University \\
\textsuperscript{2} Institute of Artificial Intelligence, Xiamen University \\
\textsuperscript{3} Key Laboratory of Digital Protection and Intelligent Processing of Intangible Cultural  \\Heritage of
Fujian and Taiwan (Xiamen University), Ministry of Culture and Tourism \\
\texttt{\{huangchongxuan,yeyongshi,biaofu,suqifeng\}@stu.xmu.edu.com} \\
\texttt{mandel@xmu.edu.com} \\
} 
\begin{document}
\maketitle
\begin{abstract}
Large language models (LLMs) have demonstrated remarkable multilingual capabilities, however, how to evaluate cross-lingual alignment remains underexplored. Existing alignment benchmarks primarily focus on sentence embeddings, but prior research has shown that neural models tend to induce a non-smooth representation space, which impact of semantic alignment evaluation on low-resource languages. 
Inspired by neuroscientific findings that similar information activates overlapping neuronal regions, we propose a novel \emph{Neuron State-Based Cross-Lingual Alignment} (\textbf{NeuronXA}) to assess the cross-lingual a lignment capabilities of LLMs, which offers a more semantically grounded approach to assess cross-lingual alignment. We evaluate NeuronXA on several prominent multilingual LLMs (LLaMA, Qwen, Mistral, GLM, and OLMo) across two transfer tasks and three multilingual benchmarks. The results demonstrate that with only 100 parallel sentence pairs, NeuronXA achieves a Pearson correlation of 0.9556 with downstream tasks performance and 0.8514 with transferability. These findings demonstrate NeuronXA's effectiveness in assessing both cross-lingual alignment and transferability, even with a small dataset. This highlights its potential to advance cross-lingual alignment research and to improve the semantic understanding of multilingual LLMs.
% and underscore its potential to drive future research on cross-lingual alignment and improve the semantic understanding of multilingual LLMs.

\end{abstract}

\section{Introduction}

\epigraph{\emph{The brain has its own language for testing the structure and consistency of the world.}}{Carl Sagan}

Recent advancements in autoregressive Large language models (LLMs) have demonstrated remarkable multilingual capabilities in understanding, reasoning, and language generation \citep{achiam2023gpt,meta2024llama3-1,yang2024qwen2}. This has spurred growing interest in evaluating their performance across diverse languages ~\citep{hendrycks2021mmlu,openai_mmmlu,ahuja-etal-2023-mega,zhang2025pmmeval,ye2025how}. However, the mechanisms underlying cross-lingual alignment in LLMs remain insufficiently understood.
% Nevertheless, the mechanisms underlying multilingual processing in LLMs remain insufficiently understood.

Research on cross-lingual alignment has focused on linguistic isomorphism in representation spaces and its impact on cross-lingual transfer~\citep{ye2023language}. Studies have explored the emergence of latent languages in multilingual processing~\citep{zhao2024llama, wendler2024llamas}, alignment dynamics during pre-training ~\citep{wang2024probing}, as well as the morphological and syntactic structures of model embeddings ~\citep{papadimitriou-etal-2021-deep}. Various strategies have been proposed to enhance alignment, including interventions at different stages of model training~\citep{Yang_Ma_Zhang_Wu_Li_Zhou_2020,zhu2024question,li2024prealign}.

% Another line of research focuses on evaluating cross-lingual alignment, particularly through the alignment of embedding spaces. 
Additionally, some research has been dedicated to evaluating cross-lingual alignment, particularly through the alignment of embedding spaces. Many studies adopt unsupervised methods to assess conceptual alignment across languages ~\citep{mousi-etal-2024-exploring}, utilizing metrics such as cosine similarity ~\citep{li2024languagerankermetricquantifying,kargaran2024mexa} to compute representational similarity. However, prior work has shown that neural architectures such as BERT and GPT tend to induce anisotropic representation spaces \citep{gaorepresentation,ethayarajh2019contextual,li2020sentence}. The collapse of representations in the semantic space diminishes the semantic expressiveness of low-resource languages ~\citep{li2024languagerankermetricquantifying}, thereby limiting the reliability of embedding-based evaluations of cross-lingual semantic alignment.

Prior studies have shown that neurons within feedforward network (FFN) modules encode diverse forms of knowledge~\cite{dai-etal-2022-knowledge, voita-etal-2023-neurons, gurnee-etal-2024-universal}. Drawing inspiration from neurobiological findings—where similar stimuli activate overlapping neural circuits—we hypothesize that neuron activations can serve as intrinsic representations of multilingual queries. These activations may provide a more structured and robust means of capturing cross-lingual knowledge, offering new insights into multilingual alignment.

In this study, we introduce a novel evaluation framework called \emph{Neuron State-Based Cross-Lingual Alignment} (\textbf{NeuronXA}) to assess the cross-lingual alignment capabilities of LLMs. The proposed method quantifies the activation likelihood of individual neurons in response to parallel corpora across multiple languages. Using neuron states as intrinsic representations, NeuronXA calculates alignment scores by evaluating the consistency of parallel sentences within the representation space, thus offering a robust method for alignment evaluation.
% By using neuron states as representations, NeuronXA calculates alignment scores by estimating the alignment of parallel sentences in the representation space, providing a robust mechanism for evaluating alignment.

Based on NeuronXA, we systematically evaluate the alignment of several popular open-source LLMs, yielding several key findings:

\begin{itemize}

\item  First, the neuron state-based representation method more effectively encodes cross-lingual knowledge. Using this intrinsic representation improves the model's accuracy in semantic retrieval, particularly in bidirectional retrieval tasks.

\item  Second, our experimental results demonstrate that the proposed NeuronXA method provides a reliable evaluation approach, exhibiting a strong correlation with both the model's transferability and its performance on multilingual benchmarks. NeuronXA offers a robust framework for assessing the cross-lingual alignment capabilities of large language models.

\item  Third, an analysis of alignment scores across different model layers reveals that the highest scores occur in the middle layers, while the lowest scores are observed in the lower and upper layers. This pattern suggests that lower layers primarily map inputs from various languages into a shared semantic space centered around high-resource languages, whereas upper layers project semantic content onto language-specific vocabulary tokens.

\end{itemize}
\section{Methods}

\subsection{Background}

Currently, LLMs are predominantly developed using the autoregressive Transformer architecture~\cite{Vaswani-etal-2017-attention}, where the core components include multi-head self-attention (MHA) and feedforward networks (FFNs). Previous research has demonstrated that the feedforward layers in Transformers can be conceptualized as key-value memory networks~\cite{geva-etal-2021-transformer}, which store world knowledge to aid in sequence understanding. Consequently, our study focuses primarily on the analysis of FFNs.

In the current LLMs architectures, FFNs typically employ gated projections for each token within a sequence. The computation for this process is defined as:
\begin{equation} \label{eq-3}
    \text{FFN}^I(\bm{x}) = \sigma\left(\bm{W_G}\bm{x} + \bm{b_G}\right) \odot \left(\bm{W^I}\bm{x} + \bm{b^I}\right),
\end{equation}
where $\bm{W_G}, \bm{W^I} \in \mathbb{R}^{d_{\text{ff}}\times d}$ and $\bm{b^I}, \bm{b_G} \in \mathbb{R}^{d_{\text{ff}}}$ represent the weight matrices and bias vectors for the input linear layer $\text{FFN}^I(\cdot)$ and the gate linear layer $\text{FFN}_G(\cdot)$, respectively. Following prior work~\citep{DBLP:conf/acl/ZhangZLXW00XS023, wang-etal-2022-finding-skill}, we can decompose the FFN layer into $d_{\text{ff}}$ neurons, each of which corresponds to a row in the input and gate layers, as well as a column in the output layer. The outputs of the FFN layers can thus be rewritten as the sum of the individual neuron outputs:
\begin{equation} \label{eq-4}
    \text{FFN}(\bm{x}) = \sum_i^{d_{\text{ff}}} \text{FFN}^I(\bm{x})_i \bm{W}^O_{:,i} + \bm{b}^O_i,
\end{equation}
where the intermediate value $\text{FFN}^I(\bm{x})_i$ denotes the activation of the $i$-th neuron.

\subsection{NeuronXA}
Previous studies have demonstrated that neurons within the FFN modules can store factual knowledge~\cite{dai-etal-2022-knowledge}, encode positional information~\cite{voita-etal-2023-neurons}, and respond to specific syntactic triggers~\cite{gurnee-etal-2024-universal}, among other functions. Building on these insights, we propose treating neuron states as intrinsic representations of the input query, with these representations potentially reflecting the various types of knowledge that underlie the query.

To capture alignment across different levels of linguistic knowledge more effectively, we leverage these neuron states as representations of the input query. Subsequently, we evaluate the alignment between queries from different languages and a high-resource-centered representation space, using this measure to define the corresponding language’s alignment score.

\paragraph{Neuron States Detection.}
\label{subsec-method-nas}
Neuron states can be detected in two distinct ways, each providing valuable insights into the model's behavior. The first method examines the neuron’s activation states, which reflect the model's response to the input. Specifically, the $j$-th neuron in the $i$-th FFN layer is considered \emph{activated} if its activation value, $\alpha(\tilde{\bm{h}}^i\bm{W}^i_1)_j$, exceeds zero~\cite{Nair-etal-2010-rectified,tang2024language}. This approach highlights the neuron’s immediate reaction to the input features.

The second method for detecting neuron partitions relies on the neuron’s absolute activation value~\label{subsec-method-nav}, which indicates the contribution of the neuron to the output of the FFN layer. This approach is commonly used as a functional indicator~\citep{DBLP:conf/acl/ZhangZLXW00XS023, wang-etal-2022-finding-skill}, where the absolute activation value of the $j$-th neuron in the $i$-th layer serves as the representation of that neuron’s role in processing a given input sentence pair.

\paragraph{Sentence Representation.}
To compute the NeuronXA score, it is first necessary to obtain the sentence representation. Unlike encoder-only models, which utilize a bidirectional attention mechanism~\citep{devlin-etal-2019-bert}, decoder-only LLMs rely on causal attention. Thus, directly averaging the representations of all tokens, as is typically done in encoder-only models, would result in an overrepresentation of early tokens, which disproportionately influences the overall sentence representation. A common approach to mitigate this issue is to use the representation of the final token \cite{neelakantan2022text,wang2023improving,ma2024fine}. However, this method does not fully capture the entire sentence. To address this limitation, \citet{muennighoff2022sgpt} proposed a position-weighted average representation, which is defined as:
\begin{equation}\label{pos-avg}
N_l = \sum_{t=1}^{T} w_t n_{lt} \quad \textrm{with} \quad w_t = \frac{t}{\sum_{k=1}^{T} k} ,
\end{equation}
where \( T \) denotes the token count of the sentence, \( n_{lt} \) represents the neuron state of the \( t \)-th token at layer \( l \), and \( N_l \) signifies the sentence neuron states at layer \( l \).

\paragraph{NeuronXA Score.}
Cross-lingual alignment refers to the tendency of semantically similar words or sentences to be closely aligned within a shared representation space ~\cite{hammerl2024understanding,kargaran2024mexa}. When the alignment between languages 
$L_1$ and $L_2$  is strong, semantically similar sentences $l_1$ and $l_2$
  should have their closest neighbors in the representation space of the opposite language. We evaluate the proportion of sentence pairs that satisfy this alignment to assess cross-lingual alignment.

We generate a square matrix \( C(l) \) representing cosine similarities of sentence representation at the output of layer \( l \) for all parallel sentences in languages \( L_1 \) and \( L_2 \).
Let \( c_{ij} \) denote the element at the \( i \)-th row and \( j \)-th column of \( C(l) \), corresponding to the cosine similarity between the \( i \)-th sentence of \( L_1 \) and the \( j \)-th sentence of \( L_2 \) at layer \( l \) of LLMs. Then we define the NeuronXA alignment score as:
\begin{equation}
\begin{split}
\mu_{C(l)} = \frac{1}{n} \sum_{i=1}^n \mathbf{1} \left( c_{ii} > \big\{ c_{ij}, c_{ji} \big\}_{j \neq i} \right),
\end{split}
\end{equation}
where \( n \) is the  the dimension of the matrix, and \( \mathbf{1}(\cdot) \) is the indicator function, which equals 1 if its argument condition evaluates to true and 0 otherwise. The calculation of this alignment score can be regarded as calculating the proportion of parallel sentences that satisfy weak alignment in the representation space.

The NeuronXA alignment score \(\mu_{ C(l)}\) is computed for language \(L_1\) with respect to pivot language \(L_2\) at each layer \(l\) of the language model. To obtain a single NeuronXA alignment score for a given the language model and language pair (\(L_1\), \(L_2\)), we use mean pooling over multiple layers.

\section{Experimental Setup}\label{sec:exp-setup}

\paragraph{Models.}
We conduct experiments on several models with approximately 7B parameters, a widely recognized baseline size in the LLM community. The models selected for evaluation include LLaMA-2, LLaMA-3, LLaMA-3.1 \cite{touvron2023llama1, dubey2024llama}, Qwen 2.5 \cite{yang2024qwen2}, Mistral 0.3 \cite{jiang2023mistral}, Olmo 2 \cite{olmo20242}, and GLM 4 \cite{glm2024chatglm}. To assess the scalability of our findings, we additionally evaluate the larger Qwen 2.5 14B model, as well as smaller models such as LLaMA-3.2 3B. These models have demonstrated strong multilingual performance and are widely adopted in the research community, making them suitable candidates for our evaluation.

\paragraph{Dataset.}
We utilize two cross-lingual parallel datasets, FLORES-200~\cite{costa2022no} and Tatoeba~\cite{artetxe2019massively}, to evaluate the effectiveness of the neuron state-based representation method in bridging the semantic gap between semantically similar sentences. To provide a more comprehensive assessment of cross-lingual alignment across a diverse set of languages, we select FLORES-200 for comparative experiments on downstream tasks, due to its extensive language coverage. A detailed discussion of the dataset can be found in Appendix \ref{app-lang_list}.

\subsection{Parallel Sentence Retrieval}

\paragraph{Problem Formulation.}
Cross-lingual parallel sentence retrieval aims to identify semantically equivalent sentences across languages, facilitating applications such as machine translation, multilingual retrieval, and cross-lingual question answering. The primary challenge is to learn sentence representations that capture meaning within a shared semantic space. The effectiveness of retrieval relies on these representations accurately preserving semantic content across different languages.

\begin{figure}[t]
    \centering
    \begin{subfigure}{0.49\linewidth}
        \includegraphics[width=\textwidth]{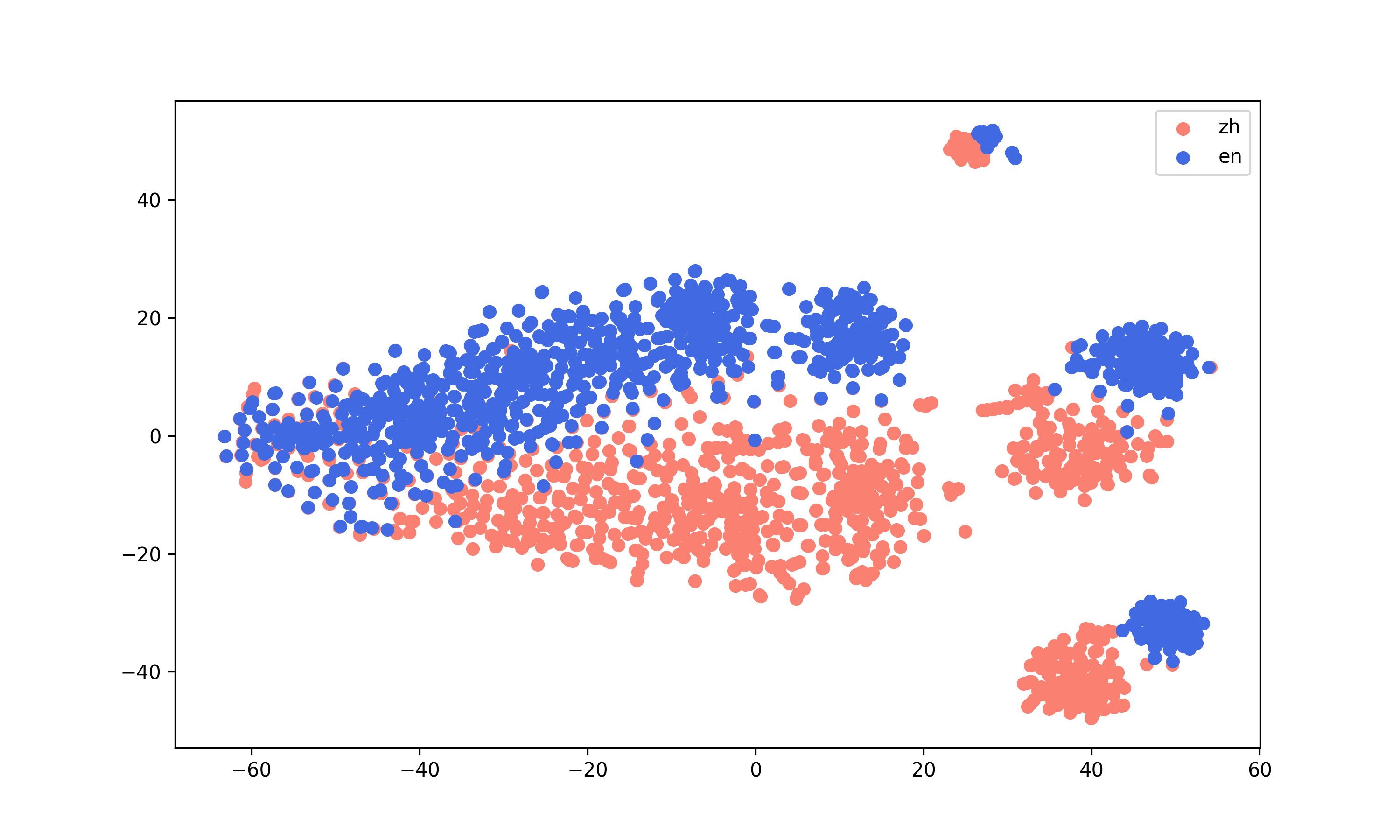}
        \caption{Sentence Embedding.}
        \label{fig:s_align}
    \end{subfigure}
    \hfill
    \begin{subfigure}{0.49\linewidth}
        \includegraphics[width=\textwidth]{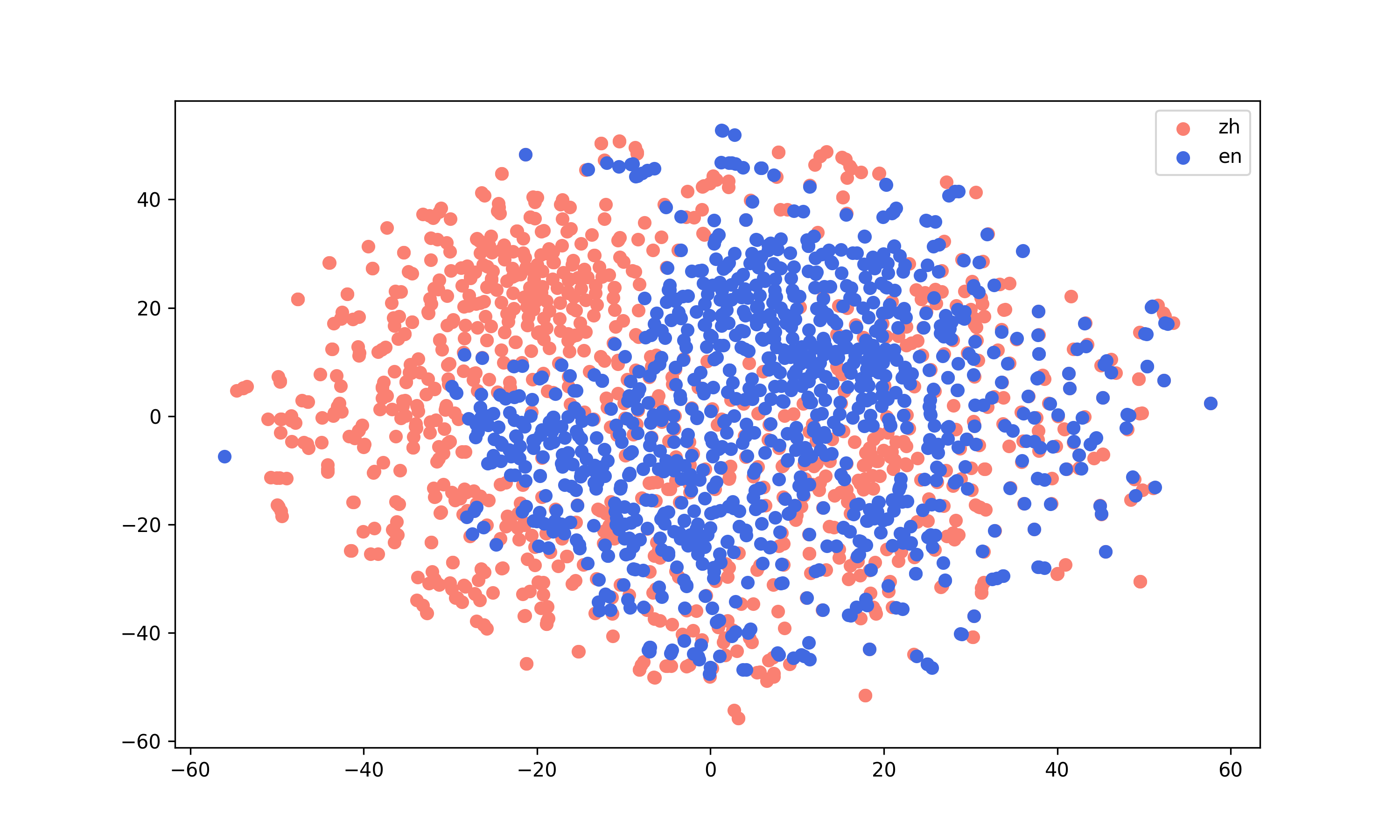}
        \caption{Neuron State.}
        \label{fig:ns_align}
    \end{subfigure}
    \caption{Visualization of sentence representations for 100 Tatoeba sentence pairs in Chinese and English, projected into 2D using t-SNE. The results compare two representation methods from Llama3.1-8B: sentence embeddings (Figure~\ref{fig:s_align}), which show significant misalignment, and the proposed NeuronXA method (Figure~\ref{fig:ns_align}), which mitigates this misalignment.}
    \label{fig:representation_compare}
\end{figure}

\paragraph{Neuron Activation-Based Representations.}
We propose \emph{Neuron Activation State (NAS)} and \emph{Neuron Activation Value (NAV)} as novel representations derived from neuron activation patterns in pre-trained language models. Unlike conventional embeddings, which suffer from issues such as non-smoothness, as illustrated in Figure \ref{fig:representation_compare}, neuron-state-based representations offer a smoother representation space, providing a more structured and interpretable approach for cross-lingual alignment.

\paragraph{Setup.}
We evaluate our method on the FLORES-200 and Tatoeba datasets, covering both head and long-tail languages (see Appendix~\ref{app-lang_list} for details). Sentence representations are constructed using a weighted token averaging strategy with the Llama 3.1-8B model. Given the model’s depth, we apply max-pooling to enhance retrieval accuracy. The primary evaluation metric is the bidirectional retrieval accuracy, which quantifies the proportion of correctly retrieved parallel sentence pairs, providing a robust assessment of representation effectiveness.

\subsection{Alignment Evaluate methods}

For comparison, we evaluate the model's cross-lingual alignment capabilities using the following methods, with assessment conducted on 100 parallel sentence pairs from the FLORES-200 dataset. The robustness of the NeuronXA method is discussed in detail in Appendix \ref{robustness}.

(a) \emph{Multilingual Evaluation via Cross-Linguistic Alignment (MEXA)}~\cite{kargaran2024mexa}: MEXA measures alignment by computing the similarity between English and non-English sentence embeddings using parallel sentences. To mitigate centralization bias, it employs relative cosine similarity for cross-lingual alignment score calculation.

(b) \emph{Neural Activation State-based Cross-Lingual Alignment (NASCA, ours)}: NASCA represents sentences based on neuron activation states (binary 0 or 1). The alignment score is derived from the proportion of parallel sentences exhibiting weak alignment in the representation space.

(c) \emph{Neural Activation Value-based Cross-Lingual Alignment (NAVCA, ours)}: NAVCA follows a similar approach to NASCA but uses the absolute magnitude of neuron activations instead of binary states. Further details are provided in Section~\ref{subsec-method-nav}.

\subsection{Cross-lingual Transfer Evaluation}
 
Following prior work \cite{li2024prealign,wang2024probing}, we assess the zero-shot cross-lingual transfer capability of models through two downstream tasks. To investigate the relationship between alignment scores and transferability, we compute the Pearson correlation coefficient between the alignment score and task performance. A higher correlation indicates that the alignment score effectively predicts the model's cross-lingual transfer ability.

\paragraph{Zero-shot Cross-lingual Transfer (ZS-CLT).}

This is a standard approach for evaluating a model’s cross-lingual generalization. In this setting, a model is fine-tuned on a given task in a source language and tested on the same task in target languages without additional training. We use the widely adopted XNLI dataset \cite{conneau-etal-2018-xnli} for evaluation, which assesses sentence understanding in multiple languages by determining the relationship between sentence pairs.

\paragraph{Cross-lingual Knowledge Application (CLKA).}
LLMs acquire extensive world knowledge from multilingual corpora. An essential capability of these models is the ability to learn knowledge in one language and apply it across others. To evaluate this ability, we use the BMLAMA-53 dataset \cite{qi2023cross}, a benchmark designed to assess cross-lingual knowledge consistency in multilingual LLMs. 

All fine-tuning experiments were conducted using the LLaMA Factory framework \cite{zheng2024llamafactory}, with prompt templates corresponding to the specific task requirements. Due to computational resource constraints, we applied 4-bit quantized LoRA \cite{hu2021lora} for fine-tuning.
\subsection{Multilingual Benchmarks Evaluation}
We evaluate model alignment by measuring how different languages are mapped into a shared representation space, which is inherently biased toward high-resource languages. As a result, alignment scores between high-resource languages and others can serve as an indirect indicator of performance in lower-resource languages. 

To evaluate this alignment, we utilize three benchmarks—Belebele \citep{bandarkar2024belebele}, m-ARC \citep{lai-etal-2023-okapi}, and m-MMLU \citep{lai-etal-2023-okapi}—which collectively encompass a diverse range of high-, medium-, and low-resource languages. 
A detailed description of these datasets is provided in Appendix~\ref{app-lang_list}.
% A detailed description of the datasets used can be found in Appendix \ref{app-lang_list}.

All experiments utilize 5-shot in-context learning via the lm-evaluation-harness framework\footnote{\href{https://github.com/EleutherAI/lm-evaluation-harness}{https://github.com/EleutherAI/lm-evaluation-harness}}, with default prompt templates for comparability.

\section{Results and Analysis}

\begin{table}[!t]
\centering
\begin{minipage}[t]{0.5\textwidth}
\makeatletter\def\@captype{table}
\footnotesize
\resizebox{\textwidth}{!}{
\begin{tabular}{llcccc}
\toprule
\multirow{2}{*}{\textbf{Representation}} & \multirow{2}{*}{\textbf{Direction}} & \multicolumn{2}{c}{\textbf{FLORES-200}} & \multicolumn{2}{c}{\textbf{Tatoeba}} 
\\
& & {Head} & {Long-tail} & {Head} & {Long-tail} \\
\midrule
\multirow{3}{*}{Embedding}
& {En $\rightarrow$ xx} & 90.12 & \color{purple}\textbf{50.71} & 23.93 & 13.28 \\
& {xx $\rightarrow$ En} & 88.60 & \color{purple}\textbf{47.74} & 54.66 & \color{purple}\textbf{41.95} \\
& {En $\Leftrightarrow$ xx} & 83.78 & 40.95 & 16.86 & 10.12 \\
\midrule
\multirow{3}{*}{NAS}
& {En $\rightarrow$ xx} & \color{purple}\textbf{93.10} & 50.49 & \color{purple}\textbf{67.77} & \color{purple}\textbf{42.12} \\
%\color{purple}\textbf{\textsc{Emma-X}} & 73.7 & 60.3 & 84.5 &  67.4 \\
& {xx $\rightarrow$ En} & \color{purple}\textbf{89.82} & 46.95 & \color{purple}\textbf{65.05} & 38.60 \\
& {En $\Leftrightarrow$ xx} & \color{purple}\textbf{87.07} & \color{purple}\textbf{42.20} & \color{purple}\textbf{57.78} & \color{purple}\textbf{32.47} \\
\bottomrule

\end{tabular}}
 \vspace{5pt}
\caption{Retrieval results on FLORES-200 and Tatoeba in xx $\rightarrow$ En and En $\rightarrow$ xx direction, along with En $\Leftrightarrow$ xx direction. The \textcolor{purple}{\textbf{bold}} font denotes the best results.}\label{tbl:long-tail-results}
\end{minipage}
\vspace{-15pt}
\end{table}

\subsection{Enhanced Semantic Alignment in Parallel Sentence Retrieval}

Table \ref{tbl:long-tail-results} shows cross-lingual semantic retrieval accuracy for different representations in the LLaMA3.1-8B model. The results highlight a performance gap between head and long-tail languages, with head languages consistently outperforming long-tail ones due to richer training data for the former, leading to stronger semantic alignment. Additionally, the impact of selecting other high-resource languages as query languages on semantic retrieval is discussed in Appendix \ref{baselang-baselines}.

\paragraph{Directional Asymmetry.} On the Tatoeba dataset, sentence embedding-based retrieval exhibits a 30.73\% accuracy drop in the En $\rightarrow$ xx direction compared to xx $\rightarrow$ En denotes a Head language. This is because English provides richer semantic representations, aiding retrieval from other languages. Conversely, when querying in English, the representation of other languages is less robust, hindering retrieval. The NAS representation, however, achieves nearly symmetric accuracy in both directions, indicating that it better captures cross-lingual semantics and mitigates representation imbalances.

\paragraph{Dataset Impact.} Retrieval accuracy is higher on FLORES-200 than on Tatoeba due to Tatoeba's lower sentence diversity, especially in low-resource languages, where semantically similar but distinct sentences complicate retrieval. In contrast, FLORES-200, sourced from Wikimedia and manually validated, offers greater diversity, enabling clearer semantic distinctions.
\paragraph{Representation Comparison.} NAS consistently outperforms sentence embeddings in bidirectional retrieval accuracy, demonstrating its superior ability to encode cross-lingual semantics as an intrinsic representation. A key advantage of NAS is its robustness in handling long-tail languages, where it achieves better alignment between high- and low-resource languages. Moreover, NAS reduces directional asymmetry, yielding nearly symmetric performance in both En $\rightarrow$ xx and xx $\rightarrow$ En retrieval tasks. This suggests that NAS provides a more balanced cross-lingual representation.

\begin{figure*}[ht]
    \centering
     \begin{subfigure}[b]{0.45\textwidth}
         \centering
         \includegraphics[width=\textwidth]{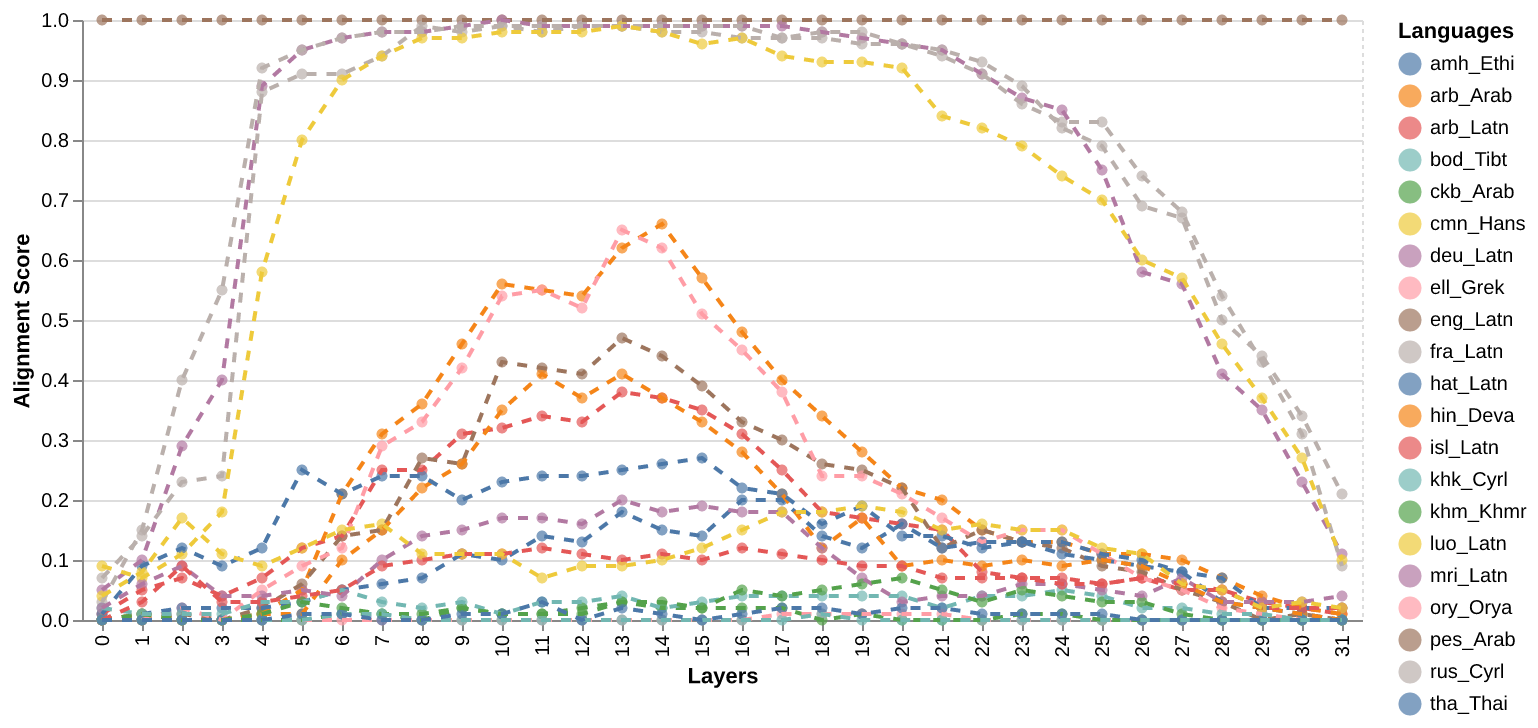}
         \caption{Llama 2 7B.}
     \end{subfigure}   
     \begin{subfigure}[b]{0.45\textwidth}
         \centering
         \includegraphics[width=\textwidth]{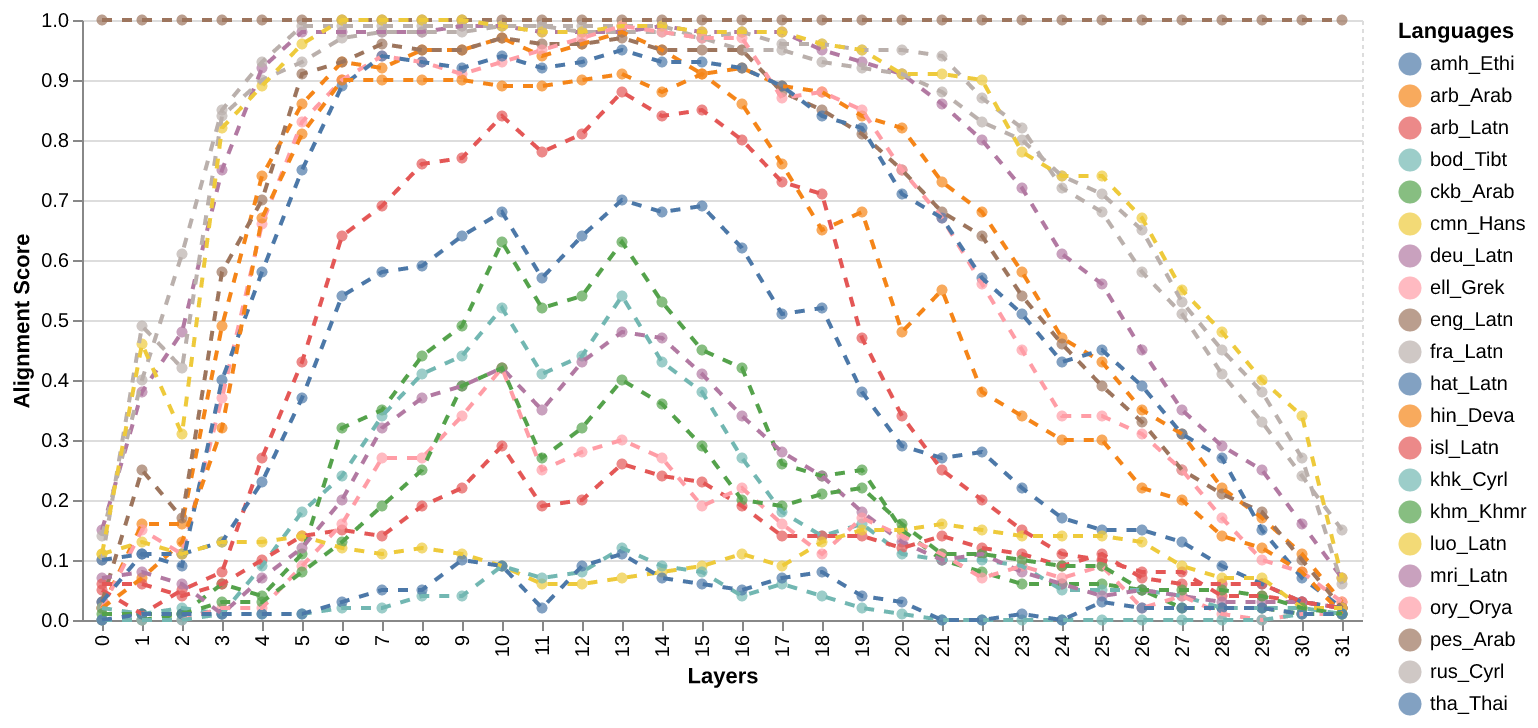}
         \caption{Llama 3.1 8B.}
     \end{subfigure}  

     %     \begin{subfigure}[b]{0.45\textwidth}
     %     \centering
     %     \includegraphics[width=\textwidth]{images/all_images/NAV-mistral-7B.png}
     %     \caption{Mistral 0.3 7B.}
     % \end{subfigure}   
     % \begin{subfigure}[b]{0.45\textwidth}
     %     \centering
     %     \includegraphics[width=\textwidth]{images/all_images/NAV-olmo-7B.png}
     %     \caption{OLMo 2 7B.}
     % \end{subfigure}   
    % \caption{Alignment scores for each model in NeuronXA default configuration for different languages across all layers.}
    \caption{NASCA scores across all layers for different languages.}
    \label{fig:body_nav_score}
\end{figure*}
% \begin{table*}[]
% \centering
% \resizebox{2\columnwidth}{!}{
% \begin{tabular}{lc|lc|lc}
% \hline
% High-High & \multicolumn{1}{l}{NASCA Score}& High-Low & \multicolumn{1}{l}{NASCA Score} & Low-Low & \multicolumn{1}{l}{NASCA Score}\\
% \hline
% English-German & 0.7300 & German-Silesian & 0.5222 & Azerbaijani-Turkmen & 0.3862 \\
% Italian-French & 0.8372 & French-Panjabi & 0.2872 & Hungarian-Yiddish & 0.3821 \\
% German-French & 0.7628 & Italian-Banjar & 0.4353 & Gujarati-Banjar & 0.2191 \\
% French-Chinese & 0.6922 & Italian-Uighur & 0.1831 & Kazakh-Tatar & 0.5291\\
% \hline
% \end{tabular}
% }
% \caption{NASCA score of different language pairs of Llama 3.1 8B.}
% \label{tabel:llama3.1 H-L}
% \end{table*}

\begin{table*}[t]
\centering
\resizebox{0.9\textwidth}{!}{
\begin{tabular}{@{}lc|lc|lc@{}}
\toprule
\multicolumn{2}{c|}{\textbf{High$\rightarrow$High}} & \multicolumn{2}{c|}{\textbf{High$\rightarrow$Low}} & \multicolumn{2}{c}{\textbf{Low$\rightarrow$Low}} \\
Language Pair & NASCA Score & Language Pair & NASCA Score & Language Pair & NASCA Score \\
\midrule
English$\rightarrow$German    & 0.7300 & German$\rightarrow$Silesian   & 0.5222 & Azerbaijani$\rightarrow$Turkmen  & 0.3862 \\
Italian$\rightarrow$French    & 0.8372 & French$\rightarrow$Panjabi    & 0.2872 & Hungarian$\rightarrow$Yiddish    & 0.3821 \\
German$\rightarrow$French     & 0.7628 & Italian$\rightarrow$Banjar    & 0.4353 & Gujarati$\rightarrow$Banjar      & 0.2191 \\
French$\rightarrow$Chinese    & 0.6922 & Italian$\rightarrow$Uighur    & 0.1831 & Kazakh$\rightarrow$Tatar         & 0.5291 \\
\bottomrule
\end{tabular}
}
\caption{NASCA score of different language pairs of Llama-3.1-8B.}
\label{tabel:llama3.1 H-L}
\end{table*}

\subsection{The Dynamics of Alignment}

\begin{table*}[ht]
\centering
\footnotesize
\resizebox{\textwidth}{!}{
\begin{tabular}{llcccccccccc}
\toprule
&  & Llama 3.1 & Llama 3 & Llama 2 & Llama 3.2 & Qwen 2.5 & Qwen 2.5 & Mistral 0.3 & OLMo2 & GLM 4 & AVG \\
 &  & 8B & 8B & 7B & 3B & 14B & 7B & 7B & 7B & 9B &  \\
\midrule
\multirow{9}{*}{\rotatebox{90}{XNLI}}
\multirow{3}{*}{\rotatebox{90}{\scalebox{.65}{weighted}}}
& MEXA & {\ul 0.9259} & 0.9211 & 0.7519 & 0.9182 & 0.6212 & 0.6898 & {\ul 0.9446                } & \color{purple}\textbf{0.9500} & 0.8107 & 0.8370 \\
 & NASCA & \color{purple}\textbf{0.9309} & {\ul 0.9271} & \color{purple}\textbf{0.9639} & {\ul 0.9401} & \color{purple}\textbf{0.8647} & \color{purple}\textbf{0.9209} & \color{purple}\textbf{0.9583} & {\ul 0.9411} & \color{purple}\textbf{0.9467} & \color{purple}\textbf{0.9326} \\
 & NAVCA & 0.9227 & \color{purple}\textbf{0.9283} & {\ul 0.9063} & \color{purple}\textbf{0.9430} & {\ul 0.8038} & {\ul 0.8415} & 0.8982 & 0.8777 & {\ul 0.9213} & {\ul 0.8937} \\

\cmidrule{2-12}
\multirow{12}{*}{\rotatebox{90}{\phantom{FLORES}}}
\multirow{3}{*}{\rotatebox{90}{\scalebox{.65}{average}}}
& MEXA & 0.7948 & 0.8379 & \color{purple}\textbf{0.7486} & 0.8506 & {\ul 0.7426} & 0.8039 & {\ul 0.9175} & \color{purple}\textbf{0.9258} & 0.8402 & {\ul 0.8291} \\
 & NASCA & \color{purple}\textbf{0.8332} & \color{purple}\textbf{0.8506} & {\ul 0.6420} & {\ul 0.8589} & \color{purple}\textbf{0.7574} & \color{purple}\textbf{0.8281} & \color{purple}\textbf{0.9373} & {\ul 0.9161} & \color{purple}\textbf{0.8575} & \color{purple}\textbf{0.8312} \\
 & NAVCA & {\ul 0.8243} & {\ul 0.8490} & 0.6253 & \color{purple}\textbf{0.8709} & 0.7270 & {\ul 0.8278} & 0.9100 & 0.8810 & {\ul 0.8570} & 0.8191 \\

\cmidrule{2-12}
\multirow{12}{*}{\rotatebox{90}{\phantom{FLORES}}}
\multirow{3}{*}{\rotatebox{90}{\scalebox{.65}{last}}}
 & MEXA & 0.7737 & 0.8023 & 0.7155 & 0.8163 & 0.5539 & 0.6010 & 0.7642 & \color{purple}\textbf{0.9377} & 0.7288 & 0.7437 \\
 & NASCA & \color{purple}\textbf{0.9250} & \color{purple}\textbf{0.9248} & \color{purple}\textbf{0.9585} & {\ul 0.9387} & \color{purple}\textbf{0.8497} & \color{purple}\textbf{0.9143} & \color{purple}\textbf{0.9547} & {\ul 0.9305} & \color{purple}\textbf{0.9422} & \color{purple}\textbf{0.9265} \\
 & NAVCA & {\ul 0.9143} & {\ul 0.9248} & {\ul 0.8959} & \color{purple}\textbf{0.9408} & {\ul 0.7964} & {\ul 0.8384} & {\ul 0.8915} & 0.8635 & {\ul 0.9181} & {\ul 0.8871} \\ 

\midrule
\midrule
\multirow{9}{*}{\rotatebox{90}{BMLAMA-53}}
\multirow{3}{*}{\rotatebox{90}{\scalebox{.65}{weighted}}}
& MEXA & 0.6567 & 0.6739 & 0.8426 & 0.6223 & 0.7522 & 0.8473 & {\ul 0.8795} & 0.7500 & \color{purple}\textbf{0.6922} & 0.7463 \\
 & NASCA & {\ul 0.6825} & {\ul 0.7187} & {\ul 0.8707} & {\ul 0.6748} & {\ul 0.7785} & {\ul 0.8575} & 0.8750 & {\ul 0.7975} & 0.6761 & {\ul 0.7701} \\
 & NAVCA & \color{purple}\textbf{0.7285} & \color{purple}\textbf{0.7415} & \color{purple}\textbf{0.8924} & \color{purple}\textbf{0.7361} & \color{purple}\textbf{0.8046} & \color{purple}\textbf{0.8773} & \color{purple}\textbf{0.9062} & \color{purple}\textbf{0.8871} & {\ul 0.6850} & \color{purple}\textbf{0.8065} \\

\cmidrule{2-12}
\multirow{12}{*}{\rotatebox{90}{\phantom{FLORES}}}
\multirow{3}{*}{\rotatebox{90}{\scalebox{.65}{average}}}
& MEXA & 0.6618 & {\ul 0.6619} & \color{purple}\textbf{0.8653} & {\ul 0.6905} & 0.7602 & {\ul 0.8312} & 0.8436 & 0.8352 & {\ul 0.7270} & {\ul 0.7641} \\
 & NASCA & \color{purple}\textbf{0.7028} & 0.6529 & {\ul 0.7739} & 0.6712 & {\ul 0.7794} & 0.8196 & {\ul 0.8462} & {\ul 0.8441} & 0.6794 & 0.7522 \\
 & NAVCA & {\ul 0.6792} & \color{purple}\textbf{0.6797} & 0.7570 & \color{purple}\textbf{0.7099} & \color{purple}\textbf{0.7909} & \color{purple}\textbf{0.8584} & \color{purple}\textbf{0.8684} & \color{purple}\textbf{0.8962} & \color{purple}\textbf{0.7290} & \color{purple}\textbf{0.7743} \\

\cmidrule{2-12}
\multirow{12}{*}{\rotatebox{90}{\phantom{FLORES}}}
\multirow{3}{*}{\rotatebox{90}{\scalebox{.65}{last}}}
& MEXA & {\ul 0.6731} & 0.6987 & 0.7826 & {\ul 0.6864} & 0.6943 & 0.6789 & 0.8378 & 0.7689 & 0.6423 & 0.7181 \\
 & NASCA & 0.6731 & {\ul 0.6999} & {\ul 0.8651} & 0.6657 & {\ul 0.7786} & {\ul 0.8537} & {\ul 0.8690} & {\ul 0.8068} & {\ul 0.6565} & {\ul 0.7632} \\
 & NAVCA & \color{purple}\textbf{0.7202} & \color{purple}\textbf{0.7304} & \color{purple}\textbf{0.8906} & \color{purple}\textbf{0.7342} & \color{purple}\textbf{0.8074} & \color{purple}\textbf{0.8688} & \color{purple}\textbf{0.9076} & \color{purple}\textbf{0.8946} & \color{purple}\textbf{0.6737} & \color{purple}\textbf{0.8031} \\

\bottomrule
\end{tabular}
}
\caption{Pearson correlation of MEXA and NeuronXA on the FLORES dataset across ZS-CLT and CLKA tasks. The values in the table represent the pearson correlation of NeuronXA and benchmark settings. The highest average correlations for each task are highlighted in \textcolor{purple}{\textbf{bold}}, and the second highest are \underline{underlined}.}
\label{pearson_clt}
\end{table*}

\begin{table*}[ht]
\centering
\footnotesize
\resizebox{1.0\linewidth}{!}{
\begin{tabular}{llcccccccccc}
\toprule
&  & Llama 3.1& Llama 3& Llama 2& Llama 3.2& Qwen 2.5 & Qwen 2.5 & Mistral 0.3& OLMo 2  & GLM 4 & AVG  \\
&  & 8B& 8B& 7B& 3B& 14B& 7B& 7B& 7B& 9B&   \\

\midrule
\multirow{9}{*}{\rotatebox{90}{m-ARC}}
\multirow{3}{*}{\rotatebox{90}{\scalebox{.65}{weighted}}}
& MEXA & 0.9551 & 0.9464 & 0.9124 & 0.9142 & 0.8709 & {\ul 0.9589} & {\ul 0.9575} & 0.8925 & 0.9225 & 0.9256 \\
 & NASCA & {\ul 0.9570} & {\ul 0.9522} & {\ul 0.9369} & {\ul 0.9186} & {\ul 0.9696} & 0.9479 & 0.9539 & {\ul 0.9177} & {\ul 0.9713} & {\ul 0.9472} \\
 & NAVCA & \color{purple}\textbf{0.9756} & \color{purple}\textbf{0.9725} & \color{purple}\textbf{0.9649} & \color{purple}\textbf{0.9522} & \color{purple}\textbf{0.9786} & \color{purple}\textbf{0.9820} & \color{purple}\textbf{0.9847} & \color{purple}\textbf{0.9569} & \color{purple}\textbf{0.9731} & \color{purple}\textbf{0.9712} \\
 
\cmidrule{2-12}
\multirow{11}{*}{\rotatebox{90}{\phantom{FLORES}}}
\multirow{3}{*}{\rotatebox{90}{\scalebox{.65}{average}}}
& MEXA & {\ul 0.9657} & \color{purple}\textbf{0.9624} & \color{purple}\textbf{0.9426} & {\ul 0.9319} & \color{purple}\textbf{0.9773} & {\ul 0.9664} & 0.9310 & 0.8800 & \color{purple}\textbf{0.9688} & {\ul 0.9473} \\
 & NASCA & 0.9650 & 0.9575 & {\ul 0.9277} & 0.9308 & {\ul 0.9692} & 0.9438 & {\ul 0.9470} & {\ul 0.8853} & 0.9592 & 0.9428 \\
 & NAVCA & \color{purple}\textbf{0.9678} & {\ul 0.9616} & 0.9241 & \color{purple}\textbf{0.9412} & 0.9686 & \color{purple}\textbf{0.9692} & \color{purple}\textbf{0.9540} & \color{purple}\textbf{0.9133} & {\ul 0.9638} & \color{purple}\textbf{0.9515} \\
 
\cmidrule{2-12}
\multirow{11}{*}{\rotatebox{90}{\phantom{FLORES}}}
\multirow{3}{*}{\rotatebox{90}{\scalebox{.65}{last}}}
& MEXA & 0.8833 & 0.8979 & 0.8853 & 0.8925 & 0.7729 & 0.7938 & 0.9279 & 0.9187 & 0.8543 & 0.8696 \\
 & NASCA & {\ul 0.9591} & {\ul 0.9535} & {\ul 0.9400} & {\ul 0.9212} & {\ul 0.9687} & {\ul 0.9510} & {\ul 0.9565} & {\ul 0.9261} & {\ul 0.9705} & {\ul 0.9496} \\
 & NAVCA & \color{purple}\textbf{0.9751} & \color{purple}\textbf{0.9728} & \color{purple}\textbf{0.9682} & \color{purple}\textbf{0.9545} & \color{purple}\textbf{0.9756} & \color{purple}\textbf{0.9804} & \color{purple}\textbf{0.9867} & \color{purple}\textbf{0.9624} & \color{purple}\textbf{0.9727} & \color{purple}\textbf{0.9720} \\

\midrule
\midrule
\multirow{9}{*}{\rotatebox{90}{m-MMLU}}
\multirow{3}{*}{\rotatebox{90}{\scalebox{.65}{weighted}}}
& MEXA & \color{purple}\textbf{0.9720} & 0.9678 & 0.9232 & 0.9543 & 0.7293 & 0.8560 & \color{purple}\textbf{0.9855} & {\ul 0.8797} & 0.8873 & 0.9061 \\
 & NASCA & {\ul 0.9704} & {\ul 0.9693} & {\ul 0.9541} & {\ul 0.9678} & \color{purple}\textbf{0.9683} & \color{purple}\textbf{0.9849} & {\ul 0.9846} & \color{purple}\textbf{0.8871} & \color{purple}\textbf{0.9717} & \color{purple}\textbf{0.9620} \\
 & NAVCA & 0.9702 & \color{purple}\textbf{0.9700} & \color{purple}\textbf{0.9762} & \color{purple}\textbf{0.9787} & {\ul 0.9322} & {\ul 0.9499} & 0.9842 & 0.8663 & {\ul 0.9673} & {\ul 0.9550} \\

\cmidrule{2-12}
\multirow{11}{*}{\rotatebox{90}{\phantom{FLORES}}}
\multirow{3}{*}{\rotatebox{90}{\scalebox{.65}{average}}}
& MEXA & {\ul 0.9638} & {\ul 0.9622} & \color{purple}\textbf{0.9300} & {\ul 0.9708} & \color{purple}\textbf{0.9170} & \color{purple}\textbf{0.9705} & {\ul 0.9698} & \color{purple}\textbf{0.9076} & \color{purple}\textbf{0.9599} & \color{purple}\textbf{0.9502} \\
 & NASCA & \color{purple}\textbf{0.9700} & \color{purple}\textbf{0.9663} & {\ul 0.8347} & \color{purple}\textbf{0.9711} & {\ul 0.9086} & {\ul 0.9696} & \color{purple}\textbf{0.9757} & {\ul 0.9035} & {\ul 0.9578} & {\ul 0.9397} \\
 & NAVCA & 0.9504 & 0.9433 & 0.8086 & 0.9652 & 0.8539 & 0.9557 & 0.9679 & 0.8802 & 0.9277 & 0.9170 \\

\cmidrule{2-12}
\multirow{12}{*}{\rotatebox{90}{\phantom{FLORES}}}
\multirow{3}{*}{\rotatebox{90}{\scalebox{.65}{last}}}
& MEXA & 0.8443 & 0.8471 & 0.8861 & 0.8448 & 0.6226 & 0.6312 & 0.8614 & {\ul 0.8772} & 0.8156 & 0.8034 \\
 & NASCA & \color{purple}\textbf{0.9675} & \color{purple}\textbf{0.9669} & {\ul 0.9574} & {\ul 0.9697} & \color{purple}\textbf{0.9597} & \color{purple}\textbf{0.9792} & \color{purple}\textbf{0.9859} & \color{purple}\textbf{0.8860} & \color{purple}\textbf{0.9661} & \color{purple}\textbf{0.9598} \\
 & NAVCA & {\ul 0.9611} & {\ul 0.9612} & \color{purple}\textbf{0.9790} & \color{purple}\textbf{0.9719} & {\ul 0.9170} & {\ul 0.9351} & {\ul 0.9783} & 0.8504 & {\ul 0.9578} & {\ul 0.9458} \\
 
\midrule
\midrule
\multirow{9}{*}{\rotatebox{90}{Belebele}}
\multirow{3}{*}{\rotatebox{90}{\scalebox{.65}{weighted}}}
& MEXA & {\ul 0.9483} & {\ul 0.9583} & 0.8108 & {\ul 0.9562} & 0.6076 & 0.7422 & {\ul 0.9745} & {\ul 0.9654} & 0.7229 & 0.8540 \\
 & NASCA & \color{purple}\textbf{0.9588} & \color{purple}\textbf{0.9614} & \color{purple}\textbf{0.9658} & \color{purple}\textbf{0.9633} & \color{purple}\textbf{0.9444} & \color{purple}\textbf{0.9494} & \color{purple}\textbf{0.9774} & \color{purple}\textbf{0.9699} & \color{purple}\textbf{0.9283} & \color{purple}\textbf{0.9576} \\
 & NAVCA & 0.9087 & 0.9214 & {\ul 0.9420} & 0.9339 & {\ul 0.8671} & {\ul 0.8501} & 0.9301 & 0.8951 & {\ul 0.8612} & {\ul 0.9011} \\

\cmidrule{2-12}
\multirow{11}{*}{\rotatebox{90}{\phantom{FLORES}}}
\multirow{3}{*}{\rotatebox{90}{\scalebox{.65}{average}}}
 & MEXA & {\ul 0.9452} & {\ul 0.9525} & \color{purple}\textbf{0.8498} & {\ul 0.9580} & {\ul 0.8572} & 0.8996 & {\ul 0.9685} & {\ul 0.9640} & 0.8888 & {\ul 0.9204} \\
 & NASCA & \color{purple}\textbf{0.9526} & \color{purple}\textbf{0.9555} & {\ul 0.7626} & \color{purple}\textbf{0.9590} & \color{purple}\textbf{0.8877} & \color{purple}\textbf{0.9416} & \color{purple}\textbf{0.9744} & \color{purple}\textbf{0.9648} & \color{purple}\textbf{0.9334} & \color{purple}\textbf{0.9257} \\
 & NAVCA & 0.9343 & 0.9387 & 0.7438 & 0.9444 & 0.8298 & {\ul 0.9016} & 0.9610 & 0.9330 & {\ul 0.9104} & 0.8997 \\

\cmidrule{2-12}
\multirow{11}{*}{\rotatebox{90}{\phantom{FLORES}}}
\multirow{3}{*}{\rotatebox{90}{\scalebox{.65}{last}}}
& MEXA & 0.6675 & 0.6907 & 0.7507 & 0.7202 & 0.4955 & 0.4924 & 0.7448 & {\ul 0.9629} & 0.5611 & 0.6762 \\
 & NASCA & \color{purple}\textbf{0.9600} & \color{purple}\textbf{0.9647} & \color{purple}\textbf{0.9621} & \color{purple}\textbf{0.9686} & \color{purple}\textbf{0.9335} & \color{purple}\textbf{0.9446} & \color{purple}\textbf{0.9796} & \color{purple}\textbf{0.9683} & \color{purple}\textbf{0.9183} & \color{purple}\textbf{0.9555} \\
 & NAVCA & {\ul 0.9089} & {\ul 0.9190} & {\ul 0.9356} & {\ul 0.9334} & {\ul 0.8569} & {\ul 0.8473} & {\ul 0.9207} & 0.8745 & {\ul 0.8508} & {\ul 0.8941}\\
\bottomrule
\end{tabular}
}
\caption{Pearson correlation of NeuronXA on the FLORES dataset across there multilingual benchmarks. The values in the table represent the correlation of NeuronXA and benchmark settings. The highest average correlations for each task are highlighted in \textcolor{purple}{\textbf{bold}}, and the second highest are \underline{underlined}.}
\label{pearson-benchmark}
\end{table*}
\paragraph{Alignment Score Across Layers.} 
Figure \ref{fig:body_nav_score} shows how alignment varies across layers, calculated using NASCA. 
As model depth increases, alignment ability initially improves and then declines, with the lowest alignment observed in both the bottom and top layers. This suggests that in generative models, neurons in the lower and upper layers are primarily language-specific, while the intermediate layers contain shared multilingual neurons, a pattern found in previous studies ~\citep{zeng-etal-2025-converging,del-fishel-2022-cross}. These findings indicate that in the lower layers, LLMs rely on language-specific neurons to map aligned text from different languages into a shared representational space for semantic transformation. In contrast, the upper layers, responsible for token generation, require a higher concentration of language-specific neurons to handle vocabulary mapping.

\paragraph{Analysis of the selection of baseline languages.}
English is selected as the pivo language for evaluating cross-lingual alignment, as LLMs often align multilingual inputs around high-resource languages. To mitigate potential biases introduced by using English as the reference, we categorize evaluation into three groups: high-resource to high-resource, high-resource to low-resource, and low-resource to low-resource. Using the FLORES-200 dataset, we select four representative language pairs for each category, with results shown in Table \ref{tabel:llama3.1 H-L}. Our analysis shows that high-resource languages exhibit relatively stable distributions, while low-resource languages show significant variability. Although English serves as a natural reference point, other high-resource languages such as German and French can also be considered as baselines.

\subsection{Downstream tasks Correlation}

In this section, we empirically evaluate the effectiveness of our proposed representation method based on neuron states. We calculate NeuronXA scores between English and other languages, and investigate their correlation with both model cross-lingual transferability and performance on multilingual tasks.

\paragraph{NeuronXA is more closely related to model transferability.}
As shown in Table \ref{pearson_clt}, both NASCA and NAVCA—our NeuronXA-based methods—outperform the sentence embedding-based baseline MEXA, which achieves an average Pearson correlation of 0.7731. In contrast, NASCA and NAVCA yield average Pearson correlations of 0.8293 and 0.8306, respectively, demonstrating a stronger correlation with the model's transferability. Notably, correlations with the CLKA task is significantly lower than that with the ZS-CLT task. We hypothesize that this gap arises from the limited size of the BMLAMA-53 dataset, which contains only 3,012 samples, potentially restricting its ability to capture real-world factual knowledge transfer. Nevertheless, both NASCA and NAVCA consistently exhibit high correlation coefficients overall.

\paragraph{NeuronXA is more closely associated with the model's multilingual capabilities.}
Similar to the results discussed in the transferability task, Table \ref{pearson-benchmark} presents the Pearson correlation coefficients between cross-lingual alignment scores and three multilingual benchmarks. The MEXA, NASCA, and NAVCA methods achieve average Pearson correlations of 0.8725, 0.9489, and 0.9341, respectively. Notably, both NASCA and NAVCA show substantial improvements in their average Pearson correlations with downstream tasks compared to MEXA.

\paragraph{Analysis of different sentence representation calculation methods.}
Token-position-based weighted sentence representation methods are generally considered to capture more contextual information, a trend reflected in both Table \ref{pearson_clt} and Table \ref{pearson-benchmark}. For both transferability tasks and multilingual benchmarks, the highest correlation coefficients are observed with the weighted method (except for the m-ARC task). The second-best performance is achieved by the average method, while the last-token method demonstrates relatively lower correlation coefficients.

Across all settings, the best overall results (higher correlation) were achieved when embeddings were computed using a weighted average and alignment scores were computed using NASCA, so we adopted this configuration as the default for NeuronXA.

Furthermore, Appendix \ref{Generative} discusses the correlation coefficient between alignment scores and generative tasks. Additionally, the robustness of NeuronXA scores when other languages serve as base languages is explored in Appendix \ref{baselang-baselines}.
\section{Related Work}

The remarkable progress in autoregressive LLMs has highlighted their exceptional multilingual competencies across comprehension, reasoning, and generative tasks ~\citep{achiam2023gpt,meta2024llama3-1,yang2024qwen2,fu2025llms,fu2025efficient}; however, the fundamental mechanisms governing these cross-linguistic capabilities remain inadequately elucidated. A systematic investigation of cross-lingual alignment through rigorous empirical evaluation could not only unravel the operational principles underlying linguistic generalization in LLMs but also inform the design of optimized methodologies for enhancing cross-lingual alignment efficiency in LLMs.

\paragraph{Multilingual mechanism.}Prior studies have demonstrated that layers closer to the model's input or output exhibit more language-specific behavior than intermediate layers ~\citep{bhattacharya-bojar-2023-unveiling}. ~\citet{zhaolarge} transformed queries into English for comprehension, conducted inference in intermediate layers using English while integrating multilingual knowledge, and generated responses consistent with the original language in the final layer. Additionally, ~\citet{wendler2024llamas} defined intermediate layers as the concept space and revealed that, for Llama models, this concept space is closer to English. Some researchers have explored the multilingual mechanisms of large models at the neuron level. ~\citet{zhang2024unveiling} found regions in large models corresponding to multilingual and monolingual capabilities. ~\citet{kojima2024multilingual} and ~\citet{bhattacharya-bojar-2023-unveiling} analyzed language-specific neurons in large models and discovered that these neurons are predominantly concentrated in the top and bottom layers of the model. Furthermore, certain studies have focused on dynamic changes. ~\citet{wang2024probing} and ~\citet{bhaskar2024heuristic} analyzed the dynamic alignment capabilities of multilingual large models during pretraining.

\paragraph{Cross-lingual Alignment.}Cross-lingual alignment can be evaluated by the similarity of representations. Several research has focused on embedding-based approaches. ~\citet{papadimitriou-etal-2021-deep} investigated morphological and syntactic alignment within embedding spaces, while ~\citet{wen-yi-mimno-2023-hyperpolyglot} studied token-level embedding similarity across models with respect to language-specific encoding patterns. ~\citet{xu-etal-2023-structural} and ~\citet{mousi-etal-2024-exploring} explored concept representation alignment in the semantic space. To evaluate cross-lingual alignment through semantic similarity, ~\citet{li2024languagerankermetricquantifying} computed cosine similarity between embeddings of parallel sentences to assess multilingual model performance. Building on this, ~\citet{kargaran2024mexa} introduced relative cosine similarity to predict alignment scores and analyzed its correlation with downstream task performance. 

Despite these advancements, the representation collapse phenomenon prevalent in neural models compromises semantic expressivity, particularly for low-resource languages ~\citep{gaorepresentation,ethayarajh2019contextual,li2020sentence}, thereby the effectiveness of embedding-based methods for cross-lingual semantic alignment is inherently limited. This limitation is also reflected in the restricted correlation with zero-shot transfer performance observed in earlier methods. Various techniques, such as Canonical Correlation Analysis ~\citep{kornblith2019similarity} and Centered Kernel Alignment ~\citep{conneau2020emerging}, have been employed to measure the similarity of intrinsic representations for parallel inputs. The work most closely related to ours is that of SADS ~\citep{zeng-etal-2025-converging}, who computed cosine similarity based on neuron activation values from parallel sentences as the cross-lingual alignment score. In contrast, our study goes further by analyzing why neuron-based approaches are effective. Furthermore, given the anisotropy issue in neural representations \citet{kargaran2024mexa}, rather than relying solely on cosine similarity values, we adopt a binary perspective. This approach ensures more reliable assessments of alignment.

\section{Conclusion}
In this paper, we propose a novel cross-lingual alignment evaluation method, Neuron State Similarity-Based Cross-Lingual Alignment (\emph{NeuronXA}), which offers a more semantically grounded approach compared to traditional methods. By leveraging NeuronXA, we assess a model's alignment ability based on the consistency of parallel sentences. Through extensive experiments, we analyze the Pearson correlation between the NeuronXA score and three downstream tasks, as well as a zero-shot cross-lingual transfer task. Our results demonstrate that the NeuronXA score is strongly correlated with both the model's transferability and its performance on multilingual tasks.

While NeuronXA demonstrates robust performance across a variety of settings, it achieves the highest alignment scores when combined with token-weighted average methods and the NASCA score evaluation approach. Notably, in the multilingual tasks, the average Pearson correlation reaches 0.9556, while the correlation with transfer tasks is 0.8514, highlighting the effectiveness of NeuronXA in capturing cross-lingual alignment. 

Overall, NeuronXA demonstrates significant potential as a robust method for evaluating the multilingual capabilities of LLMs, paving the way for future efforts to expand these models to a wider range of underrepresented languages.

\section*{Limitations}
In this study, we employ neuron states as intrinsic representations to evaluate alignment by examining the consistency of parallel sentences within the representation space. Therefore, a limitation of our evaluation method is its requirement for access to the model's intrinsic representations. Consequently, developers of closed-source models may be unable to directly apply NeuronXA. Nevertheless, they could utilize NeuronXA internally and report their results, which would provide valuable insights into their model's cross-lingual capabilities.

Moreover, various perspectives on the capabilities of large models offer alignment across different abilities. However, NeuronXA cannot encompass all of these aspects. Our goal is to provide a simple yet effective evaluation method for multilingual alignment in large models, contributing insights for future research on cross-lingual alignment and multilingual mechanisms.

\section*{Acknowledgement}
We would like to thank all the anonymous reviewers for the insightful and helpful comments.
This work is supported by National Science and Technology Major Project (Grant No. 2022ZD0116101),
the Major Scientific Research Project of the State Language Commission in the 13th Five-Year Plan (Grant No. WT135-38),  and  the public technology service platform project of Xiamen City (No. 3502Z20231043). 

\bibliography{ref}

\begin{thebibliography}{69}
\providecommand{\natexlab}[1]{#1}

\bibitem[{Ahuja et~al.(2023)Ahuja, Diddee, Hada, Ochieng, Ramesh, Jain, Nambi, Ganu, Segal, Ahmed, Bali, and Sitaram}]{ahuja-etal-2023-mega}
Kabir Ahuja, Harshita Diddee, Rishav Hada, Millicent Ochieng, Krithika Ramesh, Prachi Jain, Akshay Nambi, Tanuja Ganu, Sameer Segal, Mohamed Ahmed, Kalika Bali, and Sunayana Sitaram. 2023.
\newblock \href {https://doi.org/10.18653/v1/2023.emnlp-main.258} {{MEGA}: Multilingual evaluation of generative {AI}}.
\newblock In \emph{Proceedings of the 2023 Conference on Empirical Methods in Natural Language Processing}, pages 4232--4267, Singapore. Association for Computational Linguistics.

\bibitem[{Artetxe and Schwenk(2019)}]{artetxe2019massively}
Mikel Artetxe and Holger Schwenk. 2019.
\newblock Massively multilingual sentence embeddings for zero-shot cross-lingual transfer and beyond.
\newblock \emph{Transactions of the association for computational linguistics}, 7:597--610.

\bibitem[{Bandarkar et~al.(2024)Bandarkar, Liang, Muller, Artetxe, Shukla, Husa, Goyal, Krishnan, Zettlemoyer, and Khabsa}]{bandarkar2024belebele}
Lucas Bandarkar, Davis Liang, Benjamin Muller, Mikel Artetxe, Satya~Narayan Shukla, Donald Husa, Naman Goyal, Abhinandan Krishnan, Luke Zettlemoyer, and Madian Khabsa. 2024.
\newblock \href {https://doi.org/10.18653/v1/2024.acl-long.44} {The belebele benchmark: a parallel reading comprehension dataset in 122 language variants}.
\newblock In \emph{Proceedings of the 62nd Annual Meeting of the Association for Computational Linguistics (Volume 1: Long Papers)}, pages 749--775, Bangkok, Thailand. Association for Computational Linguistics.

\bibitem[{Bhaskar et~al.(2024)Bhaskar, Friedman, and Chen}]{bhaskar2024heuristic}
Adithya Bhaskar, Dan Friedman, and Danqi Chen. 2024.
\newblock The heuristic core: Understanding subnetwork generalization in pretrained language models.
\newblock In \emph{Proceedings of the 62nd Annual Meeting of the Association for Computational Linguistics (Volume 1: Long Papers)}, pages 14351--14368.

\bibitem[{Bhattacharya and Bojar(2023)}]{bhattacharya-bojar-2023-unveiling}
Sunit Bhattacharya and Ond{\v{r}}ej Bojar. 2023.
\newblock \href {https://doi.org/10.18653/v1/2023.blackboxnlp-1.9} {Unveiling multilinguality in transformer models: Exploring language specificity in feed-forward networks}.
\newblock In \emph{Proceedings of the 6th BlackboxNLP Workshop: Analyzing and Interpreting Neural Networks for NLP}, pages 120--126, Singapore. Association for Computational Linguistics.

\bibitem[{Clark et~al.(2018)Clark, Cowhey, Etzioni, Khot, Sabharwal, Schoenick, and Tafjord}]{clark2018think}
Peter Clark, Isaac Cowhey, Oren Etzioni, Tushar Khot, Ashish Sabharwal, Carissa Schoenick, and Oyvind Tafjord. 2018.
\newblock Think you have solved question answering? try arc, the ai2 reasoning challenge.
\newblock \emph{arXiv preprint arXiv:1803.05457}.

\bibitem[{Conneau et~al.(2018)Conneau, Rinott, Lample, Williams, Bowman, Schwenk, and Stoyanov}]{conneau-etal-2018-xnli}
Alexis Conneau, Ruty Rinott, Guillaume Lample, Adina Williams, Samuel Bowman, Holger Schwenk, and Veselin Stoyanov. 2018.
\newblock \href {https://doi.org/10.18653/v1/D18-1269} {{XNLI}: Evaluating cross-lingual sentence representations}.
\newblock In \emph{Proceedings of the 2018 Conference on Empirical Methods in Natural Language Processing}, pages 2475--2485, Brussels, Belgium. Association for Computational Linguistics.

\bibitem[{Conneau et~al.(2020)Conneau, Wu, Li, Zettlemoyer, and Stoyanov}]{conneau2020emerging}
Alexis Conneau, Shijie Wu, Haoran Li, Luke Zettlemoyer, and Veselin Stoyanov. 2020.
\newblock Emerging cross-lingual structure in pretrained language models.
\newblock In \emph{Proceedings of the 58th Annual Meeting of the Association for Computational Linguistics}, pages 6022--6034.

\bibitem[{Costa-juss{\`a} et~al.(2022)Costa-juss{\`a}, Cross, {\c{C}}elebi, Elbayad, Heafield, Heffernan, Kalbassi, Lam, Licht, Maillard et~al.}]{costa2022no}
Marta~R Costa-juss{\`a}, James Cross, Onur {\c{C}}elebi, Maha Elbayad, Kenneth Heafield, Kevin Heffernan, Elahe Kalbassi, Janice Lam, Daniel Licht, Jean Maillard, et~al. 2022.
\newblock No language left behind: Scaling human-centered machine translation.
\newblock \emph{arXiv preprint arXiv:2207.04672}.

\bibitem[{Dai et~al.(2022)Dai, Dong, Hao, Sui, Chang, and Wei}]{dai-etal-2022-knowledge}
Damai Dai, Li~Dong, Yaru Hao, Zhifang Sui, Baobao Chang, and Furu Wei. 2022.
\newblock \href {https://doi.org/10.18653/v1/2022.acl-long.581} {Knowledge neurons in pretrained transformers}.
\newblock In \emph{Proceedings of the 60th Annual Meeting of the Association for Computational Linguistics (Volume 1: Long Papers)}, pages 8493--8502, Dublin, Ireland. Association for Computational Linguistics.

\bibitem[{Del and Fishel(2022)}]{del-fishel-2022-cross}
Maksym Del and Mark Fishel. 2022.
\newblock \href {https://doi.org/10.18653/v1/2022.aacl-main.15} {Cross-lingual similarity of multilingual representations revisited}.
\newblock In \emph{Proceedings of the 2nd Conference of the Asia-Pacific Chapter of the Association for Computational Linguistics and the 12th International Joint Conference on Natural Language Processing (Volume 1: Long Papers)}, pages 185--195, Online only. Association for Computational Linguistics.

\bibitem[{Devlin et~al.(2019)Devlin, Chang, Lee, and Toutanova}]{devlin-etal-2019-bert}
Jacob Devlin, Ming-Wei Chang, Kenton Lee, and Kristina Toutanova. 2019.
\newblock \href {https://doi.org/10.18653/v1/N19-1423} {{BERT}: Pre-training of deep bidirectional transformers for language understanding}.
\newblock In \emph{Proceedings of the 2019 Conference of the North {A}merican Chapter of the Association for Computational Linguistics: Human Language Technologies, Volume 1 (Long and Short Papers)}, pages 4171--4186, Minneapolis, Minnesota. Association for Computational Linguistics.

\bibitem[{Dubey et~al.(2024{\natexlab{a}})Dubey, Jauhri, Pandey, Kadian, Al-Dahle, Letman, Mathur, Schelten, Yang, Fan et~al.}]{meta2024llama3-1}
Abhimanyu Dubey, Abhinav Jauhri, Abhinav Pandey, Abhishek Kadian, Ahmad Al-Dahle, Aiesha Letman, Akhil Mathur, Alan Schelten, Amy Yang, Angela Fan, et~al. 2024{\natexlab{a}}.
\newblock \href {https://arxiv.org/abs/2407.21783} {The {L}lama 3 herd of models}.
\newblock \emph{Preprint}, arXiv:2407.21783.

\bibitem[{Dubey et~al.(2024{\natexlab{b}})Dubey, Jauhri, Pandey, Kadian, Al-Dahle, Letman, Mathur, Schelten, Yang, Fan et~al.}]{dubey2024llama}
Abhimanyu Dubey, Abhinav Jauhri, Abhinav Pandey, Abhishek Kadian, Ahmad Al-Dahle, Aiesha Letman, Akhil Mathur, Alan Schelten, Amy Yang, Angela Fan, et~al. 2024{\natexlab{b}}.
\newblock The llama 3 herd of models.
\newblock \emph{arXiv preprint arXiv:2407.21783}.

\bibitem[{Ethayarajh(2019)}]{ethayarajh2019contextual}
Kawin Ethayarajh. 2019.
\newblock \href {https://doi.org/10.18653/v1/D19-1006} {How contextual are contextualized word representations? {C}omparing the geometry of {BERT}, {ELM}o, and {GPT}-2 embeddings}.
\newblock In \emph{Proceedings of the 2019 Conference on Empirical Methods in Natural Language Processing and the 9th International Joint Conference on Natural Language Processing (EMNLP-IJCNLP)}, pages 55--65, Hong Kong, China. Association for Computational Linguistics.

\bibitem[{Fu et~al.(2025{\natexlab{a}})Fu, Liao, Fan, Li, Zhang, Chen, and Shi}]{fu2025llms}
Biao Fu, Minpeng Liao, Kai Fan, Chengxi Li, Liang Zhang, Yidong Chen, and Xiaodong Shi. 2025{\natexlab{a}}.
\newblock \href {https://arxiv.org/abs/2504.09570} {Llms can achieve high-quality simultaneous machine translation as efficiently as offline}.
\newblock \emph{Preprint}, arXiv:2504.09570.

\bibitem[{Fu et~al.(2025{\natexlab{b}})Fu, Yu, Liao, Li, Chen, Fan, and Shi}]{fu2025efficient}
Biao Fu, Donglei Yu, Minpeng Liao, Chengxi Li, Yidong Chen, Kai Fan, and Xiaodong Shi. 2025{\natexlab{b}}.
\newblock \href {https://arxiv.org/abs/2504.11809} {Efficient and adaptive simultaneous speech translation with fully unidirectional architecture}.
\newblock \emph{Preprint}, arXiv:2504.11809.

\bibitem[{Gao et~al.(2019)Gao, He, Tan, Qin, Wang, and Liu}]{gaorepresentation}
Jun Gao, Di~He, Xu~Tan, Tao Qin, Liwei Wang, and Tieyan Liu. 2019.
\newblock \href {https://openreview.net/forum?id=SkEYojRqtm} {Representation degeneration problem in training natural language generation models}.
\newblock In \emph{International Conference on Learning Representations}.

\bibitem[{Geva et~al.(2021)Geva, Schuster, Berant, and Levy}]{geva-etal-2021-transformer}
Mor Geva, Roei Schuster, Jonathan Berant, and Omer Levy. 2021.
\newblock \href {https://doi.org/10.18653/v1/2021.emnlp-main.446} {Transformer feed-forward layers are key-value memories}.
\newblock In \emph{Proceedings of the 2021 Conference on Empirical Methods in Natural Language Processing}, pages 5484--5495, Online and Punta Cana, Dominican Republic. Association for Computational Linguistics.

\bibitem[{Gretton et~al.(2005)Gretton, Bousquet, Smola, and Sch{\"o}lkopf}]{gretton2005measuring}
Arthur Gretton, Olivier Bousquet, Alex Smola, and Bernhard Sch{\"o}lkopf. 2005.
\newblock Measuring statistical dependence with hilbert-schmidt norms.
\newblock In \emph{International conference on algorithmic learning theory}, pages 63--77. Springer.

\bibitem[{Guo et~al.(2024)Guo, Yang, Li, Wei, Shang, and Chen}]{guo2024novel}
Jiaxin Guo, Hao Yang, Zongyao Li, Daimeng Wei, Hengchao Shang, and Xiaoyu Chen. 2024.
\newblock A novel paradigm boosting translation capabilities of large language models.
\newblock In \emph{Findings of the Association for Computational Linguistics: NAACL 2024}, pages 639--649.

\bibitem[{Gurnee et~al.(2024)Gurnee, Horsley, Guo, Kheirkhah, Sun, Hathaway, Nanda, and Bertsimas}]{gurnee-etal-2024-universal}
Wes Gurnee, Theo Horsley, Zifan~Carl Guo, Tara~Rezaei Kheirkhah, Qinyi Sun, Will Hathaway, Neel Nanda, and Dimitris Bertsimas. 2024.
\newblock Universal neurons in gpt2 language models.
\newblock \emph{CoRR}.

\bibitem[{H{\"a}mmerl et~al.(2024)H{\"a}mmerl, Libovick{\`y}, and Fraser}]{hammerl2024understanding}
Katharina H{\"a}mmerl, Jind{\v{r}}ich Libovick{\`y}, and Alexander Fraser. 2024.
\newblock Understanding cross-lingual alignment—a survey.
\newblock In \emph{Findings of the Association for Computational Linguistics ACL 2024}, pages 10922--10943.

\bibitem[{Hendrycks et~al.(2021{\natexlab{a}})Hendrycks, Burns, Basart, Zou, Mazeika, Song, and Steinhardt}]{hendrycks2021mmlu}
Dan Hendrycks, Collin Burns, Steven Basart, Andy Zou, Mantas Mazeika, Dawn Song, and Jacob Steinhardt. 2021{\natexlab{a}}.
\newblock Measuring massive multitask language understanding.
\newblock In \emph{International Conference on Learning Representations}.

\bibitem[{Hendrycks et~al.(2021{\natexlab{b}})Hendrycks, Burns, Basart, Zou, Mazeika, Song, and Steinhardt}]{openai_mmmlu}
Dan Hendrycks, Collin Burns, Steven Basart, Andy Zou, Mantas Mazeika, Dawn Song, and Jacob Steinhardt. 2021{\natexlab{b}}.
\newblock Measuring massive multitask language understanding.
\newblock In \emph{International Conference on Learning Representations}.

\bibitem[{Hu et~al.(2022)Hu, Wallis, Allen-Zhu, Li, Wang, Wang, Chen et~al.}]{hu2021lora}
Edward~J Hu, Phillip Wallis, Zeyuan Allen-Zhu, Yuanzhi Li, Shean Wang, Lu~Wang, Weizhu Chen, et~al. 2022.
\newblock Lora: Low-rank adaptation of large language models.
\newblock In \emph{International Conference on Learning Representations}.

\bibitem[{Hu et~al.(2020)Hu, Ruder, Siddhant, Neubig, Firat, and Johnson}]{pmlr-v119-hu20b}
Junjie Hu, Sebastian Ruder, Aditya Siddhant, Graham Neubig, Orhan Firat, and Melvin Johnson. 2020.
\newblock \href {https://proceedings.mlr.press/v119/hu20b.html} {{XTREME}: A massively multilingual multi-task benchmark for evaluating cross-lingual generalisation}.
\newblock In \emph{Proceedings of the 37th International Conference on Machine Learning}, volume 119 of \emph{Proceedings of Machine Learning Research}, pages 4411--4421. PMLR.

\bibitem[{Jiang et~al.(2023)Jiang, Sablayrolles, Mensch, Bamford, Chaplot, Casas, Bressand, Lengyel, Lample, Saulnier et~al.}]{jiang2023mistral}
Albert~Q Jiang, Alexandre Sablayrolles, Arthur Mensch, Chris Bamford, Devendra~Singh Chaplot, Diego de~las Casas, Florian Bressand, Gianna Lengyel, Guillaume Lample, Lucile Saulnier, et~al. 2023.
\newblock Mistral 7b.
\newblock \emph{arXiv preprint arXiv:2310.06825}.

\bibitem[{Kargaran et~al.(2024)Kargaran, Modarressi, Nikeghbal, Diesner, Yvon, and Sch{\"u}tze}]{kargaran2024mexa}
Amir~Hossein Kargaran, Ali Modarressi, Nafiseh Nikeghbal, Jana Diesner, Fran{\c{c}}ois Yvon, and Hinrich Sch{\"u}tze. 2024.
\newblock Mexa: Multilingual evaluation of english-centric llms via cross-lingual alignment.
\newblock \emph{arXiv preprint arXiv:2410.05873}.

\bibitem[{Kojima et~al.(2024)Kojima, Okimura, Iwasawa, Yanaka, and Matsuo}]{kojima2024multilingual}
Takeshi Kojima, Itsuki Okimura, Yusuke Iwasawa, Hitomi Yanaka, and Yutaka Matsuo. 2024.
\newblock On the multilingual ability of decoder-based pre-trained language models: Finding and controlling language-specific neurons.
\newblock In \emph{Proceedings of the 2024 Conference of the North American Chapter of the Association for Computational Linguistics: Human Language Technologies (Volume 1: Long Papers)}, pages 6919--6971.

\bibitem[{Kornblith et~al.(2019)Kornblith, Norouzi, Lee, and Hinton}]{kornblith2019similarity}
Simon Kornblith, Mohammad Norouzi, Honglak Lee, and Geoffrey Hinton. 2019.
\newblock Similarity of neural network representations revisited.
\newblock In \emph{International conference on machine learning}, pages 3519--3529. PMLR.

\bibitem[{Lai et~al.(2023)Lai, Nguyen, Ngo, Nguyen, Dernoncourt, Rossi, and Nguyen}]{lai-etal-2023-okapi}
Viet Lai, Chien Nguyen, Nghia Ngo, Thuat Nguyen, Franck Dernoncourt, Ryan Rossi, and Thien Nguyen. 2023.
\newblock \href {https://doi.org/10.18653/v1/2023.emnlp-demo.28} {Okapi: Instruction-tuned large language models in multiple languages with reinforcement learning from human feedback}.
\newblock In \emph{Proceedings of the 2023 Conference on Empirical Methods in Natural Language Processing: System Demonstrations}, pages 318--327, Singapore. Association for Computational Linguistics.

\bibitem[{Li et~al.(2020)Li, Zhou, He, Wang, Yang, and Li}]{li2020sentence}
Bohan Li, Hao Zhou, Junxian He, Mingxuan Wang, Yiming Yang, and Lei Li. 2020.
\newblock On the sentence embeddings from pre-trained language models.
\newblock In \emph{Proceedings of the 2020 Conference on Empirical Methods in Natural Language Processing (EMNLP)}, pages 9119--9130.

\bibitem[{Li et~al.(2024)Li, Huang, Ching, Dai, and Chen}]{li2024prealign}
Jiahuan Li, Shujian Huang, Aarron Ching, Xinyu Dai, and Jiajun Chen. 2024.
\newblock Prealign: Boosting cross-lingual transfer by early establishment of multilingual alignment.
\newblock In \emph{Proceedings of the 2024 Conference on Empirical Methods in Natural Language Processing}, pages 10246--10257.

\bibitem[{Li et~al.(2025)Li, Shi, Liu, Yang, Payani, Liu, and Du}]{li2024languagerankermetricquantifying}
Zihao Li, Yucheng Shi, Zirui Liu, Fan Yang, Ali Payani, Ninghao Liu, and Mengnan Du. 2025.
\newblock Language ranker: A metric for quantifying llm performance across high and low-resource languages.
\newblock In \emph{Proceedings of the AAAI Conference on Artificial Intelligence}, volume~39, pages 28186--28194.

\bibitem[{Ma et~al.(2024)Ma, Wang, Yang, Wei, and Lin}]{ma2024fine}
Xueguang Ma, Liang Wang, Nan Yang, Furu Wei, and Jimmy Lin. 2024.
\newblock Fine-tuning llama for multi-stage text retrieval.
\newblock In \emph{Proceedings of the 47th International ACM SIGIR Conference on Research and Development in Information Retrieval}, pages 2421--2425.

\bibitem[{Mousi et~al.(2024)Mousi, Durrani, Dalvi, Hawasly, and Abdelali}]{mousi-etal-2024-exploring}
Basel Mousi, Nadir Durrani, Fahim Dalvi, Majd Hawasly, and Ahmed Abdelali. 2024.
\newblock \href {https://doi.org/10.18653/v1/2024.acl-long.344} {Exploring alignment in shared cross-lingual spaces}.
\newblock In \emph{Proceedings of the 62nd Annual Meeting of the Association for Computational Linguistics (Volume 1: Long Papers)}, pages 6326--6348, Bangkok, Thailand. Association for Computational Linguistics.

\bibitem[{Muennighoff(2022)}]{muennighoff2022sgpt}
Niklas Muennighoff. 2022.
\newblock {SGPT}: {GPT} sentence embeddings for semantic search.
\newblock \emph{arXiv preprint arXiv:2202.08904}.

\bibitem[{Nair and Hinton(2010)}]{Nair-etal-2010-rectified}
Vinod Nair and Geoffrey~E. Hinton. 2010.
\newblock Rectified linear units improve restricted boltzmann machines.
\newblock In \emph{Proceedings of the 27th International Conference on International Conference on Machine Learning}, ICML'10, page 807–814, Madison, WI, USA. Omnipress.

\bibitem[{Neelakantan et~al.(2022)Neelakantan, Xu, Puri, Radford, Han, Tworek, Yuan, Tezak, Kim, Hallacy, Heidecke, Shyam, Power, Nekoul, Sastry, Krueger, Schnurr, Such, Hsu, Thompson, Khan, Sherbakov, Jang, Welinder, and Weng}]{neelakantan2022text}
Arvind Neelakantan, Tao Xu, Raul Puri, Alec Radford, Jesse~Michael Han, Jerry Tworek, Qiming Yuan, Nikolas Tezak, Jong~Wook Kim, Chris Hallacy, Johannes Heidecke, Pranav Shyam, Boris Power, Tyna~Eloundou Nekoul, Girish Sastry, Gretchen Krueger, David Schnurr, Felipe~Petroski Such, Kenny Hsu, Madeleine Thompson, Tabarak Khan, Toki Sherbakov, Joanne Jang, Peter Welinder, and Lilian Weng. 2022.
\newblock \href {https://arxiv.org/abs/2201.10005} {Text and code embeddings by contrastive pre-training}.
\newblock \emph{Preprint}, arXiv:2201.10005.

\bibitem[{OLMo et~al.(2024)OLMo, Walsh, Soldaini, Groeneveld, Lo, Arora, Bhagia, Gu, Huang, Jordan et~al.}]{olmo20242}
Team OLMo, Pete Walsh, Luca Soldaini, Dirk Groeneveld, Kyle Lo, Shane Arora, Akshita Bhagia, Yuling Gu, Shengyi Huang, Matt Jordan, et~al. 2024.
\newblock 2 olmo 2 furious.
\newblock \emph{arXiv preprint arXiv:2501.00656}.

\bibitem[{OpenAI et~al.(2023)OpenAI, Achiam, Adler, Agarwal, Ahmad, Akkaya, Aleman, Almeida, Altenschmidt, Altman, Anadkat et~al.}]{achiam2023gpt}
OpenAI, Josh Achiam, Steven Adler, Sandhini Agarwal, Lama Ahmad, Ilge Akkaya, Florencia~Leoni Aleman, Diogo Almeida, Janko Altenschmidt, Sam Altman, Shyamal Anadkat, et~al. 2023.
\newblock {GPT}-4 technical report.
\newblock \emph{arXiv preprint arXiv:2303.08774}.

\bibitem[{Papadimitriou et~al.(2021)Papadimitriou, Chi, Futrell, and Mahowald}]{papadimitriou-etal-2021-deep}
Isabel Papadimitriou, Ethan~A. Chi, Richard Futrell, and Kyle Mahowald. 2021.
\newblock \href {https://doi.org/10.18653/v1/2021.eacl-main.215} {Deep subjecthood: Higher-order grammatical features in multilingual {BERT}}.
\newblock In \emph{Proceedings of the 16th Conference of the European Chapter of the Association for Computational Linguistics: Main Volume}, pages 2522--2532, Online. Association for Computational Linguistics.

\bibitem[{Qi et~al.(2023)Qi, Fern{\'a}ndez, and Bisazza}]{qi2023cross}
Jirui Qi, Raquel Fern{\'a}ndez, and Arianna Bisazza. 2023.
\newblock Cross-lingual consistency of factual knowledge in multilingual language models.
\newblock In \emph{The 2023 Conference on Empirical Methods in Natural Language Processing}.

\bibitem[{Raghu et~al.(2017)Raghu, Gilmer, Yosinski, and Sohl-Dickstein}]{raghu2017svcca}
Maithra Raghu, Justin Gilmer, Jason Yosinski, and Jascha Sohl-Dickstein. 2017.
\newblock Svcca: Singular vector canonical correlation analysis for deep learning dynamics and interpretability.
\newblock \emph{Advances in neural information processing systems}, 30.

\bibitem[{Tang et~al.(2024)Tang, Luo, Huang, Zhang, Wang, Zhao, Wei, and Wen}]{tang2024language}
Tianyi Tang, Wenyang Luo, Haoyang Huang, Dongdong Zhang, Xiaolei Wang, Wayne~Xin Zhao, Furu Wei, and Ji-Rong Wen. 2024.
\newblock Language-specific neurons: The key to multilingual capabilities in large language models.
\newblock In \emph{Proceedings of the 62nd Annual Meeting of the Association for Computational Linguistics (Volume 1: Long Papers)}, pages 5701--5715.

\bibitem[{Touvron et~al.(2023)Touvron, Lavril, Izacard, Martinet, Lachaux, Lacroix, Rozi{\`e}re, Goyal, Hambro, Azhar et~al.}]{touvron2023llama1}
Hugo Touvron, Thibaut Lavril, Gautier Izacard, Xavier Martinet, Marie-Anne Lachaux, Timoth{\'e}e Lacroix, Baptiste Rozi{\`e}re, Naman Goyal, Eric Hambro, Faisal Azhar, et~al. 2023.
\newblock Llama: Open and efficient foundation language models.
\newblock \emph{arXiv preprint arXiv:2302.13971}.

\bibitem[{Vaswani et~al.(2017)Vaswani, Shazeer, Parmar, Uszkoreit, Jones, Gomez, Kaiser, and Polosukhin}]{Vaswani-etal-2017-attention}
Ashish Vaswani, Noam Shazeer, Niki Parmar, Jakob Uszkoreit, Llion Jones, Aidan~N Gomez, \L~ukasz Kaiser, and Illia Polosukhin. 2017.
\newblock \href {https://proceedings.neurips.cc/paper_files/paper/2017/file/3f5ee243547dee91fbd053c1c4a845aa-Paper.pdf} {Attention is all you need}.
\newblock In \emph{Advances in Neural Information Processing Systems}, volume~30. Curran Associates, Inc.

\bibitem[{Voita et~al.(2024)Voita, Ferrando, and Nalmpantis}]{voita-etal-2023-neurons}
Elena Voita, Javier Ferrando, and Christoforos Nalmpantis. 2024.
\newblock Neurons in large language models: Dead, n-gram, positional.
\newblock In \emph{Findings of the Association for Computational Linguistics ACL 2024}, pages 1288--1301.

\bibitem[{Wang et~al.(2024{\natexlab{a}})Wang, Minervini, and Ponti}]{wang2024probing}
Hetong Wang, Pasquale Minervini, and Edoardo Ponti. 2024{\natexlab{a}}.
\newblock Probing the emergence of cross-lingual alignment during llm training.
\newblock In \emph{Findings of the Association for Computational Linguistics ACL 2024}, pages 12159--12173.

\bibitem[{Wang et~al.(2024{\natexlab{b}})Wang, Yang, Huang, Yang, Majumder, and Wei}]{wang2023improving}
Liang Wang, Nan Yang, Xiaolong Huang, Linjun Yang, Rangan Majumder, and Furu Wei. 2024{\natexlab{b}}.
\newblock Improving text embeddings with large language models.
\newblock In \emph{Proceedings of the 62nd Annual Meeting of the Association for Computational Linguistics (Volume 1: Long Papers)}, pages 11897--11916.

\bibitem[{Wang et~al.(2022)Wang, Wen, Zhang, Hou, Liu, and Li}]{wang-etal-2022-finding-skill}
Xiaozhi Wang, Kaiyue Wen, Zhengyan Zhang, Lei Hou, Zhiyuan Liu, and Juanzi Li. 2022.
\newblock \href {https://doi.org/10.18653/v1/2022.emnlp-main.765} {Finding skill neurons in pre-trained transformer-based language models}.
\newblock In \emph{Proceedings of the 2022 Conference on Empirical Methods in Natural Language Processing}, pages 11132--11152, Abu Dhabi, United Arab Emirates. Association for Computational Linguistics.

\bibitem[{Wen-Yi and Mimno(2023)}]{wen-yi-mimno-2023-hyperpolyglot}
Andrea~W Wen-Yi and David Mimno. 2023.
\newblock \href {https://doi.org/10.18653/v1/2023.emnlp-main.71} {Hyperpolyglot {LLM}s: Cross-lingual interpretability in token embeddings}.
\newblock In \emph{Proceedings of the 2023 Conference on Empirical Methods in Natural Language Processing}, pages 1124--1131, Singapore. Association for Computational Linguistics.

\bibitem[{Wendler et~al.(2024)Wendler, Veselovsky, Monea, and West}]{wendler2024llamas}
Chris Wendler, Veniamin Veselovsky, Giovanni Monea, and Robert West. 2024.
\newblock Do llamas work in english? on the latent language of multilingual transformers.
\newblock In \emph{Proceedings of the 62nd Annual Meeting of the Association for Computational Linguistics (Volume 1: Long Papers)}, pages 15366--15394.

\bibitem[{Xu et~al.(2023{\natexlab{a}})Xu, Kim, Sharaf, and Awadalla}]{xu2023paradigm}
Haoran Xu, Young~Jin Kim, Amr Sharaf, and Hany~Hassan Awadalla. 2023{\natexlab{a}}.
\newblock A paradigm shift in machine translation: Boosting translation performance of large language models.
\newblock In \emph{The Twelfth International Conference on Learning Representations}.

\bibitem[{Xu et~al.(2023{\natexlab{b}})Xu, Zhang, Ye, Zhang, and Huang}]{xu-etal-2023-structural}
Ningyu Xu, Qi~Zhang, Jingting Ye, Menghan Zhang, and Xuanjing Huang. 2023{\natexlab{b}}.
\newblock \href {https://doi.org/10.18653/v1/2023.findings-emnlp.931} {Are structural concepts universal in transformer language models? towards interpretable cross-lingual generalization}.
\newblock In \emph{Findings of the Association for Computational Linguistics: EMNLP 2023}, pages 13951--13976, Singapore. Association for Computational Linguistics.

\bibitem[{Yang et~al.(2024)Yang, Yang, Zhang, Hui, Zheng, Yu, Li, Liu, Huang, Wei et~al.}]{yang2024qwen2}
An~Yang, Baosong Yang, Beichen Zhang, Binyuan Hui, Bo~Zheng, Bowen Yu, Chengyuan Li, Dayiheng Liu, Fei Huang, Haoran Wei, et~al. 2024.
\newblock Qwen2. 5 technical report.
\newblock \emph{arXiv preprint arXiv:2412.15115}.

\bibitem[{Yang et~al.(2020)Yang, Ma, Zhang, Wu, Li, and Zhou}]{Yang_Ma_Zhang_Wu_Li_Zhou_2020}
Jian Yang, Shuming Ma, Dongdong Zhang, ShuangZhi Wu, Zhoujun Li, and Ming Zhou. 2020.
\newblock \href {https://doi.org/10.1609/aaai.v34i05.6480} {Alternating language modeling for cross-lingual pre-training}.
\newblock \emph{Proceedings of the AAAI Conference on Artificial Intelligence}, 34(05):9386--9393.

\bibitem[{Ye et~al.(2023)Ye, Tao, and Kong}]{ye2023language}
Jiacheng Ye, Xijia Tao, and Lingpeng Kong. 2023.
\newblock Language versatilists vs. specialists: An empirical revisiting on multilingual transfer ability.
\newblock \emph{arXiv preprint arXiv:2306.06688}.

\bibitem[{Ye et~al.(2025)Ye, Fu, Huang, Chen, and Shi}]{ye2025how}
Yongshi Ye, Biao Fu, Chongxuan Huang, Yidong Chen, and Xiaodong Shi. 2025.
\newblock \href {https://arxiv.org/abs/2505.19987} {How well do large reasoning models translate? a comprehensive evaluation for multi-domain machine translation}.
\newblock \emph{Preprint}, arXiv:2505.19987.

\bibitem[{Zeng et~al.(2024)Zeng, Xu, Wang, Zhang, Yin, Rojas, Feng, Zhao, Lai, Yu et~al.}]{glm2024chatglm}
Aohan Zeng, Bin Xu, Bowen Wang, Chenhui Zhang, Da~Yin, Diego Rojas, Guanyu Feng, Hanlin Zhao, Hanyu Lai, Hao Yu, et~al. 2024.
\newblock Chatglm: A family of large language models from glm-130b to glm-4 all tools.
\newblock \emph{CoRR}.

\bibitem[{Zeng et~al.(2025)Zeng, Han, Chen, and Yu}]{zeng-etal-2025-converging}
Hongchuan Zeng, Senyu Han, Lu~Chen, and Kai Yu. 2025.
\newblock \href {https://aclanthology.org/2025.coling-main.707/} {Converging to a lingua franca: Evolution of linguistic regions and semantics alignment in multilingual large language models}.
\newblock In \emph{Proceedings of the 31st International Conference on Computational Linguistics}, pages 10602--10617, Abu Dhabi, UAE. Association for Computational Linguistics.

\bibitem[{Zhang et~al.(2025)Zhang, Wan, Deng, Yang, Wei, Huang, Yu, Lin, Huang, and Zhou}]{zhang2025pmmeval}
Yidan Zhang, Yu~Wan, Boyi Deng, Baosong Yang, Haoran Wei, Fei Huang, Bowen Yu, Junyang Lin, Fei Huang, and Jingren Zhou. 2025.
\newblock \href {https://arxiv.org/abs/2411.09116} {P-mmeval: A parallel multilingual multitask benchmark for consistent evaluation of llms}.
\newblock \emph{Preprint}, arXiv:2411.09116.

\bibitem[{Zhang et~al.(2023)Zhang, Zeng, Lin, Xiao, Wang, Han, Liu, Xie, Sun, and Zhou}]{DBLP:conf/acl/ZhangZLXW00XS023}
Zhengyan Zhang, Zhiyuan Zeng, Yankai Lin, Chaojun Xiao, Xiaozhi Wang, Xu~Han, Zhiyuan Liu, Ruobing Xie, Maosong Sun, and Jie Zhou. 2023.
\newblock Emergent modularity in pre-trained transformers.
\newblock In \emph{Proceedings of ACL: Findings}, pages 4066--4083.

\bibitem[{Zhang et~al.(2024)Zhang, Zhao, Zhang, Gui, and Huang}]{zhang2024unveiling}
Zhihao Zhang, Jun Zhao, Qi~Zhang, Tao Gui, and Xuan-Jing Huang. 2024.
\newblock Unveiling linguistic regions in large language models.
\newblock In \emph{Proceedings of the 62nd Annual Meeting of the Association for Computational Linguistics (Volume 1: Long Papers)}, pages 6228--6247.

\bibitem[{Zhao et~al.(2024{\natexlab{a}})Zhao, Zhang, Zhang, Gui, and Huang}]{zhao2024llama}
Jun Zhao, Zhihao Zhang, Qi~Zhang, Tao Gui, and Xuanjing Huang. 2024{\natexlab{a}}.
\newblock Llama beyond english: An empirical study on language capability transfer.
\newblock \emph{arXiv preprint arXiv:2401.01055}.

\bibitem[{Zhao et~al.(2024{\natexlab{b}})Zhao, Zhang, Chen, Kawaguchi, and Bing}]{zhaolarge}
Yiran Zhao, Wenxuan Zhang, Guizhen Chen, Kenji Kawaguchi, and Lidong Bing. 2024{\natexlab{b}}.
\newblock How do large language models handle multilingualism?
\newblock In \emph{The Thirty-eighth Annual Conference on Neural Information Processing Systems}.

\bibitem[{Zheng et~al.(2024)Zheng, Zhang, Zhang, Ye, Luo, Feng, and Ma}]{zheng2024llamafactory}
Yaowei Zheng, Richong Zhang, Junhao Zhang, Yanhan Ye, Zheyan Luo, Zhangchi Feng, and Yongqiang Ma. 2024.
\newblock \href {http://arxiv.org/abs/2403.13372} {Llamafactory: Unified efficient fine-tuning of 100+ language models}.
\newblock In \emph{Proceedings of the 62nd Annual Meeting of the Association for Computational Linguistics (Volume 3: System Demonstrations)}, Bangkok, Thailand. Association for Computational Linguistics.

\bibitem[{Zhu et~al.(2024)Zhu, Huang, Yuan, She, Chen, and Birch}]{zhu2024question}
Wenhao Zhu, Shujian Huang, Fei Yuan, Shuaijie She, Jiajun Chen, and Alexandra Birch. 2024.
\newblock Question translation training for better multilingual reasoning.
\newblock In \emph{Findings of the Association for Computational Linguistics ACL 2024}, pages 8411--8423.

\end{thebibliography}

\clearpage
\appendix
\section{Dataset}
\label{app-lang_list}
\subsection{Parallel Datasets}
\paragraph{FLORES-200.} This multilingual parallel corpus consists of English sentences sampled in equal proportions from Wikinews, Wikijunior, and Wikivoyage. Each sentence has been translated into more than 200 languages, with data quality ensured through a combination of automated validation and human review. Since the test set is not publicly available, our experiments are conducted on the dev-test set, which consists of 1,012 sentences covering 213 languages. We set the 68 languages: {``bel\_Cyrl, bos\_Latn, hun\_Latn, epo\_Latn, khm\_Khmr, urd\_Arab, srp\_Cyrl, jav\_Latn, hye\_Armn, gla\_Latn, por\_Latn, lit\_Latn, bul\_Cyrl, slk\_Latn, mal\_Mlym, ita\_Latn, nno\_Latn, mar\_Deva, hrv\_Latn, hin\_Deva, kat\_Geor, ben\_Beng, fin\_Latn, cym\_Latn, oci\_Latn, cat\_Latn, fao\_Latn, xho\_Latn, spa\_Latn, ron\_Latn, amh\_Ethi, ces\_Latn, swe\_Latn, nld\_Latn, tat\_Cyrl, kor\_Hang, glg\_Latn, fra\_Latn, eus\_Latn, ind\_Latn, dan\_Latn, tha\_Thai, deu\_Latn, tel\_Telu, afr\_Latn, pol\_Latn, est\_Latn, uig\_Arab, ukr\_Cyrl, uzn\_Latn, heb\_Hebr, kaz\_Cyrl, nob\_Latn, rus\_Cyrl, vie\_Latn, arb\_Arab, zho\_Hans, tuk\_Latn, khk\_Cyrl, jpn\_Jpan, ell\_Grek, isl\_Latn, tam\_Taml, slv\_Latn, tur\_Latn, mkd\_Cyrl, tgl\_Latn, gle\_Latn''} as ``Head'' languages, and the remaining 135 languages (excluded English data) as ``Long-tail'' ones.

\paragraph{Tatoeba.} The Tatoeba dataset\cite{artetxe2019massively} serves as a benchmark for evaluating multilingual sentence embeddings in similarity search tasks. It covers 112 languages and provides up to 1,000 English-aligned sentence pairs for each language. The evaluation is performed by computing cosine similarity to retrieve the nearest neighbors of each sentence in other languages, followed by calculating the error rate. We treat the 36 languages contained in XTREME\cite{pmlr-v119-hu20b} as head languages, which are: “ar, he, vi, id, jv, tl, eu, ml, ta, te, af, nl, en, de, el, bn, hi, mr, ur, fa, fr, it, pt, es, bg, ru, ja, ka, ko, th, sw, zh, kk, tr, et, fi, hu, az, lt, pl, uk, ro”. The remaining 76 languages in Tatoeba are treated as long-tail ones.
\subsection{Multilingual Benchmarks}

\paragraph{Belebele.} A multilingual multiple-choice machine reading comprehension dataset spanning 122 language variants. It evaluates both monolingual and multilingual models across resource-rich and resource-scarce languages. Each item consists of a question, four answer choices, and a passage sourced from FLORES-200. The dataset is meticulously annotated to distinguish proficiency levels, with rigorous quality control measures. Since five languages in Belebele are not present in FLORES-200, our analysis focuses on the 117 overlapping languages.

\paragraph{m-ARC.} The Multilingual AI2 Reasoning Challenge extends the original English ARC benchmark\cite{clark2018think} to assess cross-lingual scientific reasoning. It consists of systematically translated multiple-choice questions in 31 languages, generated using GPT-3.5-Turbo. The dataset includes 1,116 training items, 1,169 test items, and 298 validation items, all aligned with scientific reasoning objectives and grade-school science curricula.

\paragraph{m-MMLU.} A multilingual extension of the MMLU benchmark \citep{hendrycks2021mmlu}, covering 34 languages. The dataset was initially translated into 31 languages using GPT-3.5-Turbo, with expert translations for Icelandic and Norwegian. It contains 277 training items, 13,258 test items, and 1,433 validation items, spanning four domains: humanities, social sciences, STEM disciplines, and professional subjects. As the most comprehensive multilingual knowledge benchmark, m-MMLU provides a robust evaluation of cross-lingual understanding.

\begin{figure*}[h]
    \centering
         \begin{subfigure}[b]{0.4\textwidth}
         \centering
         \includegraphics[width=\textwidth]{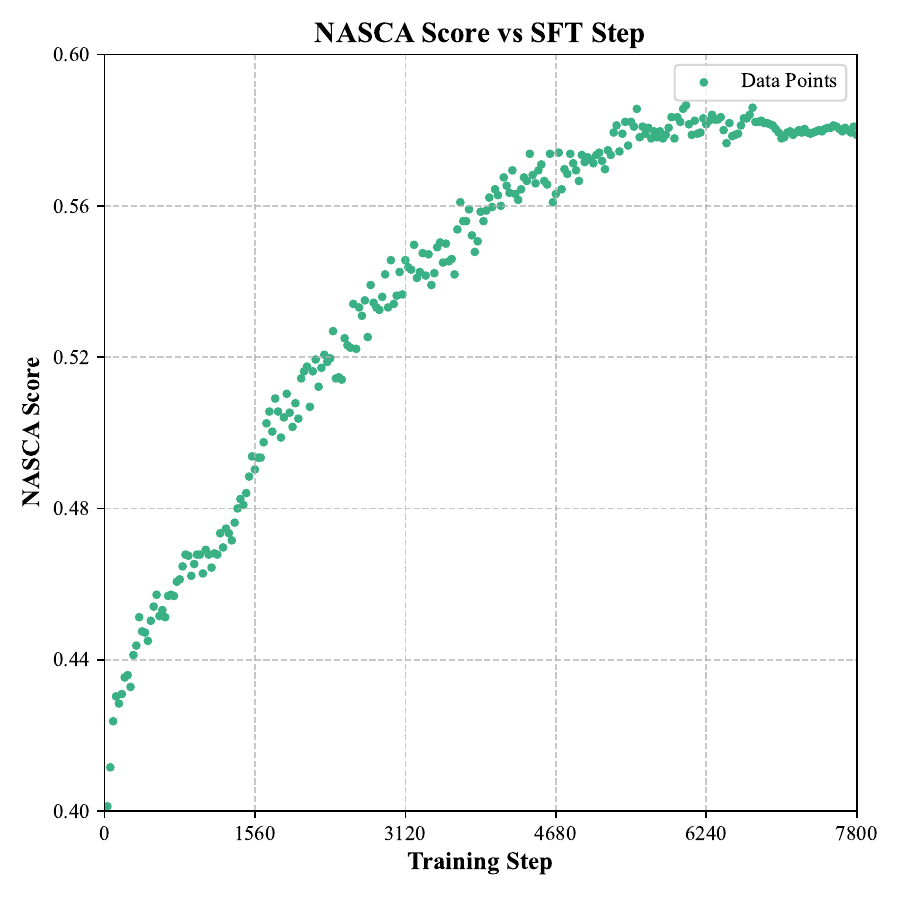}
         \caption{Supervised fine-tuning.}
     \end{subfigure}   
     \begin{subfigure}[b]{0.4\textwidth}
         \centering
         \includegraphics[width=\textwidth]{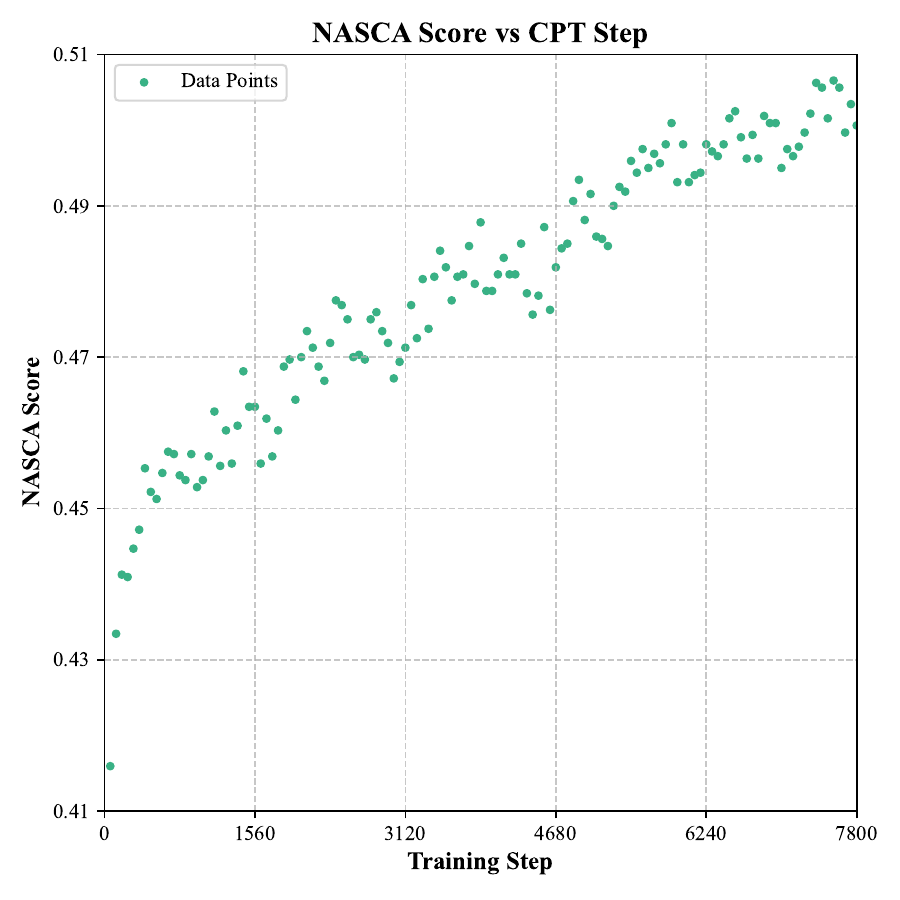}
         \caption{Continue Pre-training.}
     \end{subfigure} 
    \caption{Alignment Score Trends During Supervised Fine-Tuning and Continued Pre-Training of LLaMA-3.1 8B.}
    \label{fig:app_bleu}
\end{figure*}

\begin{figure*}[h]
    \centering
         \begin{subfigure}[b]{0.4\textwidth}
         \centering
         \includegraphics[width=\textwidth]{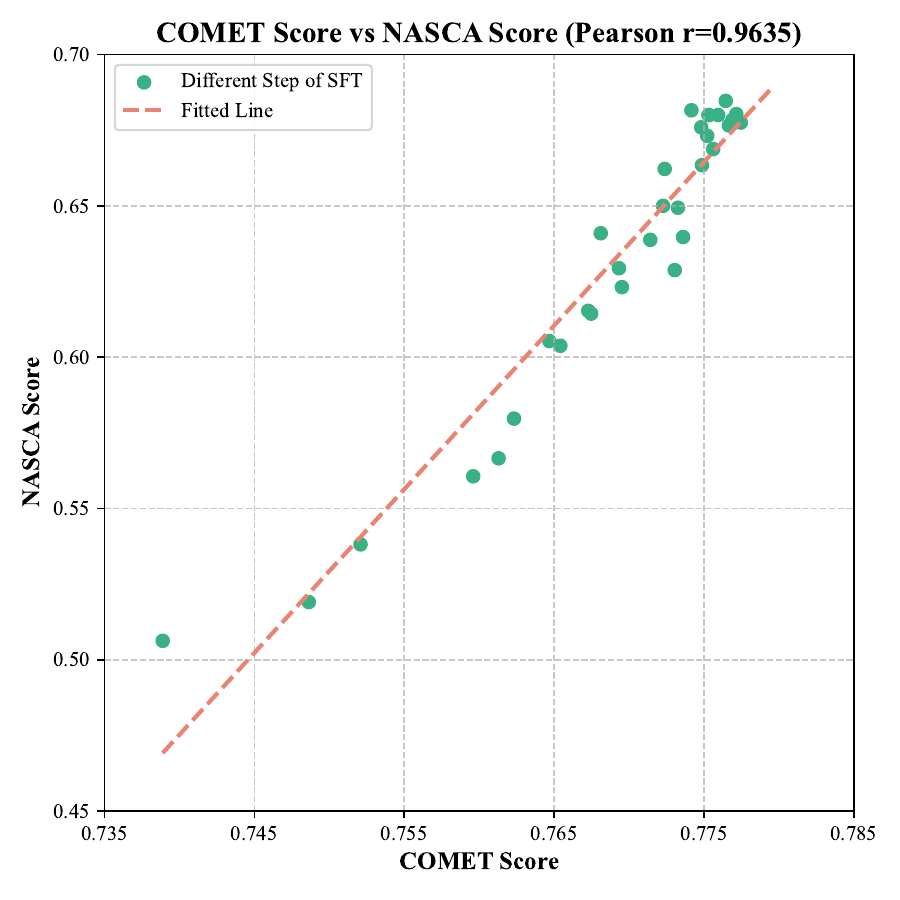}
         \caption{COMET.}
     \end{subfigure}   
     \begin{subfigure}[b]{0.4\textwidth}
         \centering
         \includegraphics[width=\textwidth]{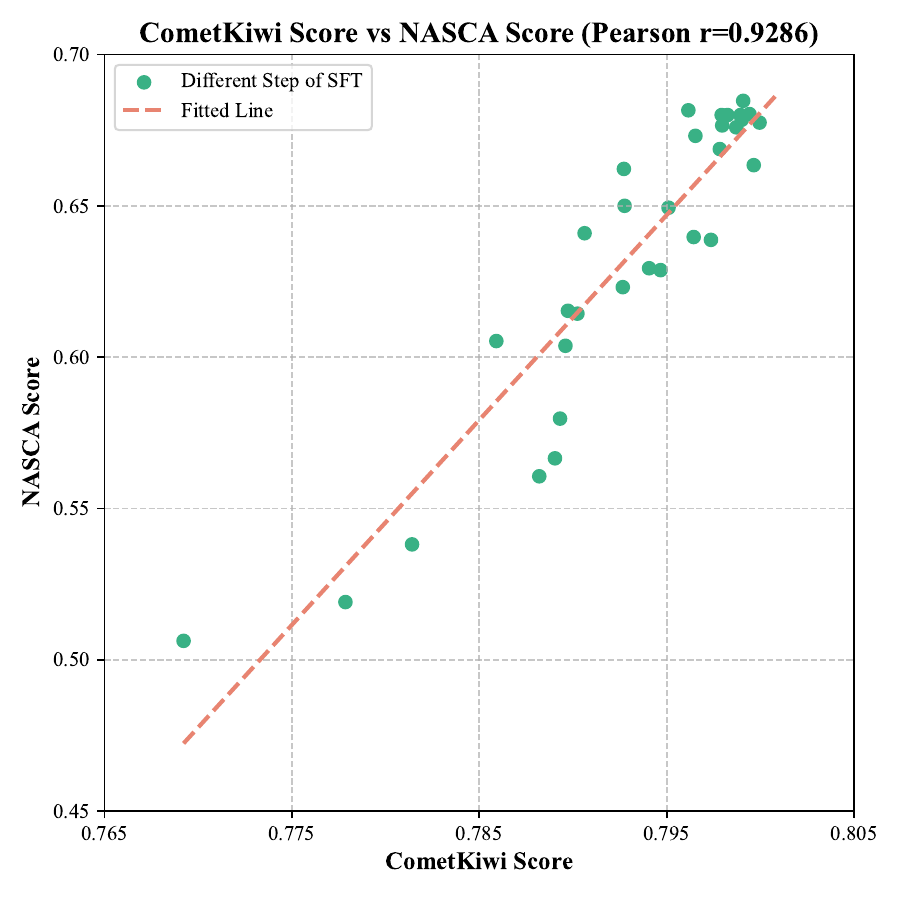}
         \caption{CometKiwi.}
     \end{subfigure} 
    \caption{Correlation coefficients between Alignment Scores and COMET/CometKiwi Scores during Supervised Fine-Tuning of LLaMA-3.1 8B.}
    \label{app:correlation}
\end{figure*}

\section{Generative Tasks Evaluation}
\label{Generative}
Certain generation tasks are strongly correlated with a model's cross-lingual alignment capabilities. In the context of machine translation, several training paradigms have been proposed to enhance a model's ability to map low-resource languages into a unified representation space with high-resource languages \cite{xu2023paradigm, guo2024novel}. These approaches aim to improve the model’s understanding of low-resource languages, fostering emergent multilingual alignment during fine-tuning.

Given this, we hypothesize that a model's translation performance is closely related to its alignment ability. To test this, we selected the NLLB \citep{costa2022no} dataset, specifically 1 million sentence pairs of English and Icelandic (with a 1:1 ratio), and fine-tuned the model with 4-bit quantized LoRA for supervised training. We used the NAS to assess alignment at each fine-tuning step and calculated the Pearson correlation between the alignment scores and the COMET/CometKiwi scores at each step.

\paragraph{Fine-tuning facilitates the alignment of the model to a unified representation space.} As shown in Figure \ref{fig:app_bleu}, the alignment scores increase with fluctuations during the fine-tuning process, indicating that fine-tuning promotes the alignment of languages in the training data into a shared space.

\paragraph{NeuronXA is closely related to machine translation performance.}
In our analysis, we computed the Pearson correlation between COMET/CometKiwi scores and alignment scores at each step, resulting in a correlation coefficient of 0.9635 and 0.9286, respectively. This strong correlation indicates that alignment scores are highly indicative of translation performance. Furthermore, alignment serves as a valuable metric for evaluating the model's translation capabilities.

\section{Other Baselines}
\label{baselang-baselines}

\begin{table*}[h]
\centering
\begin{minipage}[t]{0.9\textwidth}
\makeatletter\def\@captype{table}
\footnotesize
\resizebox{\textwidth}{!}{
\begin{tabular}{llcccccc}
\toprule
\multirow{2}{*}{\textbf{Base Language}} & \multirow{2}{*}{\textbf{Representation}} & \multicolumn{3}{c}{\textbf{Head}} & \multicolumn{3}{c}{\textbf{Long-tail}} 
\\
& & {src $\rightarrow$ xx} & {xx $\rightarrow$ src} & {src $\Leftrightarrow$ xx} & {src $\rightarrow$ xx} & {xx $\rightarrow$ src} & {src $\Leftrightarrow$ xx} \\
\midrule
\multirow{2}{*}{French} & Embedding & 86.31 & 81.16 & 78.53 & 49.69 & 41.14 & 38.74 \\
& NAS & \color{purple}\textbf{93.35} & \color{purple}\textbf{88.29} & \color{purple}\textbf{86.03} & \color{purple}\textbf{51.57} & \color{purple}\textbf{45.91} & \color{purple}\textbf{42.02} \\
\midrule
\multirow{2}{*}{Italian} & Embedding & 85.96 & 81.99 & 78.77 & 49.30 & 42.27 & 39.15 \\
& NAS & \color{purple}\textbf{92.48} & \color{purple}\textbf{88.60} & \color{purple}\textbf{85.84} & \color{purple}\textbf{50.92} & \color{purple}\textbf{46.71} & \color{purple}\textbf{41.89} \\
\midrule
\multirow{2}{*}{German} & Embedding & 86.69 & 82.51 & 79.48 & 50.16 & 43.15 & 40.05 \\
 & NAS & \color{purple}\textbf{94.40} & \color{purple}\textbf{89.00} & \color{purple}\textbf{87.01} & \color{purple}\textbf{52.53} & \color{purple}\textbf{46.54} & \color{purple}\textbf{42.48} \\
\bottomrule

\end{tabular}}
\caption{Retrieval results on FLORES-200 in xx $\rightarrow$ src and src $\rightarrow$ xx direction, along with src $\Leftrightarrow$ xx direction. The bold font denotes the best results.}\label{tapp:other-retrieval}
\end{minipage}
\end{table*}

% \begin{table*}[ht]
% \centering
% \begin{minipage}[t]{0.65\textwidth}
% \makeatletter\def\@captype{table}
% \footnotesize
% \resizebox{\textwidth}{!}{
% \begin{tabular}{lcccc}
% \toprule
% Representation & English & German & French & Italian \\
% \midrule
% MEXA & 0.9585 & 0.9250 & 0.9389 & 0.9356 \\
% NAS & \color{purple}\textbf{0.9621} & \color{purple}\textbf{0.9287} & \color{purple}\textbf{0.9460} & \color{purple}\textbf{0.9428} \\
% \bottomrule

% \end{tabular}}
% \caption{Average Pearson correlation of MEXA and NeuronXA across marc, mmlu and belebele tasks.}\label{tapp:downstream-task}
% \end{minipage}
% \end{table*}

\begin{table}[h]
\centering
\begin{minipage}[t]{0.48\textwidth}
\makeatletter\def\@captype{table}
\footnotesize
\resizebox{\textwidth}{!}{
\begin{tabular}{lcccc}
\toprule
\multirow{2}{*}{\textbf{Baselines}} & \multicolumn{4}{c}{\textbf{Languages}} \\
 & English & German & French & Italian \\
\midrule
MEXA & 0.9585 & 0.9250 & 0.9389 & 0.9356 \\ 
NASCA & \color{purple}\textbf{0.9621} & \color{purple}\textbf{0.9287} & \color{purple}\textbf{0.9460} & \color{purple}\textbf{0.9428} \\
\bottomrule
\end{tabular}}
\caption{Average Pearson correlation of MEXA and NeuronXA across marc, mmlu and belebele tasks.}\label{tapp:downstream-task}
\end{minipage}
\end{table}

\begin{table}[h]
\centering
\begin{minipage}[t]{0.48\textwidth}
\makeatletter\def\@captype{table}
\footnotesize
\resizebox{\textwidth}{!}{
\begin{tabular}{llccc}
\toprule
\multirow{2}{*}{\textbf{Base Language}} & \multirow{2}{*}{\textbf{Baselines}} & \multirow{2}{*}{\textbf{XNLI}} & \multirow{2}{*}{\textbf{Bmlama}} & \multirow{2}{*}{\textbf{Average}} \\ \\
\midrule
\multirow{2}{*}{English} & MEXA & 0.9259 & 0.6567 & 0.7913 \\
& NASCA & \color{purple}\textbf{0.9309} & \color{purple}\textbf{0.6825} & \color{purple}\textbf{0.8067} \\
\midrule
\multirow{2}{*}{German} & MEXA & \color{purple}\textbf{0.8993} & 0.5613 & 0.7303 \\
 & NASCA & 0.8986 & \color{purple}\textbf{0.6656} & \color{purple}\textbf{0.7821} \\
\bottomrule

\end{tabular}}
\caption{Average Pearson correlation of MEXA and NeuronXA across XNLI, Bmlama tasks.}\label{tapp:transfer}
\end{minipage}
\end{table}
\subsection{Exploring Other Languages as Base Languages}

In the domain of multilingual modeling, English was selected as the primary base language for this study due to its predominant role in mainstream multilingual models. Nevertheless, we acknowledge the importance of evaluating the generalizability of our method to other high-resource languages. To this end, we conducted a series of experiments, including semantic retrieval, downstream task performance correlation, and cross-lingual transferability correlation.

Specifically, for cross-lingual retrieval and downstream task correlation experiments, we employed German, French, and Italian as base languages on the LLaMA-3.1 8B model. The transferability correlation experiments were conducted using German as the base language.

As shown in Table~\ref{tapp:other-retrieval}, when using these high-resource languages for semantic retrieval, the NAS-based method consistently outperformed the embedding-based approach in retrieval accuracy across all three languages. These results align with our English-based findings, suggesting that the NAS-based method generalizes well to other base languages.

We further computed NASCA scores using German, French, and Italian as base languages and evaluated their correlation with downstream task performance. As presented in Table~\ref{tapp:downstream-task}, the NASCA scores maintained strong correlations even with non-English base languages.

Finally, we assessed the relationship between alignment scores and cross-lingual transferability using German. The results, reported in Table~\ref{tapp:transfer}, further confirm the robustness and cross-linguistic applicability of our approach.

\subsection{Other Baselines}
\label{other-baselines}
\begin{table*}[ht]
\centering
% \footnotesize
\resizebox{1.0\textwidth}{!}{
\begin{tabular}{llcccccccc}
\toprule
&  & Llama 3.1& Llama 2& Llama 3.2 & Qwen 2.5 & Mistral 0.3& OLMo 2  & GLM 4 & AVG  \\
&  & 8B& 7B& 3B& 7B& 7B& 7B& 9B&   \\

\midrule
\multirow{6}{*}{\rotatebox{90}{m-ARC}}
& CKA & 0.8333 & 0.8328 & 0.7711 & 0.7989 & 0.8774 & 0.8109 & 0.8955 & 0.8314 \\
 & SVCCA & 0.9303 & 0.9189 & 0.8865 & 0.9172 & 0.9329 & 0.8519 & 0.9370 & 0.9107 \\
 & ANC & {\ul 0.9683} & {\ul 0.9385} & {\ul 0.9305} & {\ul 0.9633} & {\ul 0.9659} & 0.9025 & 0.9690 & {\ul 0.9483} \\
 & MEXA & 0.9551 & 0.9124 & 0.9142 & 0.9589 & 0.9575 & 0.8925 & 0.9225 & 0.9304 \\
 &NASCA & 0.9570 & 0.9369 & 0.9186 & 0.9479 & 0.9539 & {\ul 0.9177} & {\ul 0.9713} & 0.9433 \\
 &NAVCA & \color{purple}\textbf{0.9756} & \color{purple}\textbf{0.9649} & \color{purple}\textbf{0.9522} & \color{purple}\textbf{0.9820} & \color{purple}\textbf{0.9847} & \color{purple}\textbf{0.9569} & \color{purple}\textbf{0.9731} & \color{purple}\textbf{0.9699} \\
\midrule
\midrule
\multirow{6}{*}{\rotatebox{90}{m-MMLU}}
& CKA & 0.8779 & 0.8642 & 0.8468 & 0.8836 & 0.9256 & 0.8745 & 0.9451 & 0.8882 \\
 & SVCCA & 0.9478 & 0.9691 & 0.9454 & 0.9202 & 0.9693 & \color{purple}\textbf{0.8936} & 0.9474 & 0.9418 \\
 & ANC & 0.9522 & {\ul 0.9760} & 0.9619 & 0.9051 & 0.9770 & {\ul 0.8907} & 0.9618 & 0.9464 \\
 & MEXA & \color{purple}\textbf{0.9720} & 0.9232 & 0.9543 & 0.8560 & \color{purple}\textbf{0.9855} & 0.8797 & 0.8873 & 0.9226 \\
 &NASCA & {\ul 0.9704} & 0.9541 & {\ul 0.9678} & \color{purple}\textbf{0.9849} & {\ul 0.9846} & 0.8871 & \color{purple}\textbf{0.9717} & \color{purple}\textbf{0.9601} \\
 &NAVCA & 0.9702 & \color{purple}\textbf{0.9762} & \color{purple}\textbf{0.9787} & {\ul 0.9499} & 0.9842 & 0.8663 & {\ul 0.9673} & {\ul 0.9561} \\
\midrule
\midrule
\multirow{6}{*}{\rotatebox{90}{Belebele}}
& CKA & 0.4751 & 0.5824 & 0.5239 & 0.3463 & 0.6317 & 0.9260 & 0.4293 & 0.5592 \\
 & SVCCA & 0.9157 & {\ul 0.9530} & 0.9354 & {\ul 0.8732} & 0.9431 & 0.9439 & {\ul 0.8858} & {\ul 0.9214} \\
 & ANC & 0.9022 & 0.9378 & 0.9331 & 0.8324 & 0.9313 & 0.9257 & 0.8319 & 0.8992 \\
 & MEXA & {\ul 0.9483} & 0.8108 & {\ul 0.9562} & 0.7422 & {\ul 0.9745} & {\ul 0.9654} & 0.7229 & 0.8743 \\
 &NASCA & \color{purple}\textbf{0.9588} & \color{purple}\textbf{0.9658} & \color{purple}\textbf{0.9633} & \color{purple}\textbf{0.9494} & \color{purple}\textbf{0.9774} & \color{purple}\textbf{0.9699} & \color{purple}\textbf{0.9283} & \color{purple}\textbf{0.9590} \\
 &NAVCA & 0.9087 & 0.9420 & 0.9339 & 0.8501 & 0.9301 & 0.8951 & 0.8612 & 0.9030 \\
\bottomrule
\end{tabular}
}
\caption{Pearson correlation of NeuronXA on the FLORES dataset across there multilingual benchmarks. The values in the table represent the correlation of NeuronXA and benchmark. The highest average correlations for each task are highlighted in \textcolor{purple}{\textbf{bold}}, and the second highest are \underline{underlined}.}
\label{app-baselines}
\end{table*}

\begin{table*}[ht]
\centering
% \footnotesize
\resizebox{1.0\linewidth}{!}{
\begin{tabular}{llcccccccc}
\toprule
&  & Llama 3.1& Llama 2& Llama 3.2 & Qwen 2.5 & Mistral 0.3& OLMo 2  & GLM 4 & AVG  \\
&  & 8B& 7B& 3B& 7B& 7B& 7B& 9B&   \\

\midrule
\multirow{6}{*}{\rotatebox{90}{XNLI}}
& CKA & 0.6694 & 0.6612 & 0.6879 & 0.4022 & 0.6604 & 0.8944 & 0.6601 & 0.6622 \\
 & SVCCA & 0.8800 & 0.8846 & 0.9144 & 0.7610 & 0.8897 & 0.8714 & 0.8797 & 0.8687 \\
 & ANC & 0.8645 & 0.8815 & 0.9082 & 0.7751 & 0.8784 & 0.8534 & 0.8775 & 0.8627 \\
 & MEXA & {\ul 0.9259} & 0.7519 & 0.9182 & 0.6898 & {\ul 0.9446} & \color{purple}\textbf{0.9500} & 0.8107 & 0.8559 \\
 &NASCA & \color{purple}\textbf{0.9309} & \color{purple}\textbf{0.9639} & {\ul 0.9401} & \color{purple}\textbf{0.9209} & \color{purple}\textbf{0.9583} & {\ul 0.9411} & \color{purple}\textbf{0.9467} & \color{purple}\textbf{0.9431} \\
 &NAVCA & 0.9227 & {\ul 0.9063} & \color{purple}\textbf{0.9430} & {\ul 0.8415} & 0.8982 & 0.8777 & {\ul 0.9213} & {\ul 0.9015} \\
\midrule
\midrule
\multirow{6}{*}{\rotatebox{90}{BMLAMA}}
& CKA & 0.6694 & 0.8280 & 0.5313 & 0.7855 & 0.6893 & 0.7196 & 0.6662 & 0.6985 \\
 & SVCCA & \color{purple}\textbf{0.8800} & 0.8516 & 0.7039 & 0.8608 & 0.8247 & 0.8234 & {\ul 0.7441} & 0.8126 \\
 & ANC & {\ul 0.8645} & {\ul 0.8895} & {\ul 0.7239} & {\ul 0.8663} & 0.8646 & \color{purple}\textbf{0.9156} & \color{purple}\textbf{0.7891} & \color{purple}\textbf{0.8448} \\
 & MEXA & 0.6567 & 0.8426 & 0.6223 & 0.8473 & {\ul 0.8795} & 0.7500 & 0.6922 & 0.7558 \\
 &NASCA & 0.6825 & 0.8707 & 0.6748 & 0.8575 & 0.8750 & 0.7975 & 0.6761 & 0.7763 \\
 &NAVCA & 0.7285 & \color{purple}\textbf{0.8924} & \color{purple}\textbf{0.7361} & \color{purple}\textbf{0.8773} & \color{purple}\textbf{0.9062} & {\ul 0.8871} & 0.6850 & {\ul 0.8161} \\
\bottomrule
\end{tabular}
}
\caption{Pearson correlation of NeuronXA on the FLORES dataset across ZS-CLT and CLKA tasks. The values in the table represent the correlation of benchmarks. The highest average correlations for each task are highlighted in \textcolor{purple}{\textbf{bold}}, and the second highest are \underline{underlined}.}
\label{app-baselines-transfer}
\end{table*}

\begin{table*}[h]
\centering
\begin{minipage}[t]{0.6\textwidth}
\makeatletter\def\@captype{table}
\footnotesize
\resizebox{\textwidth}{!}{
\begin{tabular}{lcc}
\toprule
\multirow{1}{*}{\textbf{Baselines}} & \multirow{1}{*}{\textbf{Multilingual performance}} & \multirow{1}{*}{\textbf{Cross-lingual transferability}} \\
\midrule
CKA & 0.7596 & 0.6804 \\
SVCCA & 0.9246 & 0.8407 \\
ANC & 0.9313 & 0.8537 \\
MEXA & 0.9091 & 0.8058 \\
\rowcolor[gray]{0.95} \multicolumn{3}{c}{{\textit{ours}}}\\ 
NASCA & \color{purple}\textbf{0.9541} & \color{purple}\textbf{0.8597} \\
NAVCA & 0.9430 & 0.8588 \\
\bottomrule

\end{tabular}}
\caption{Average Pearson correlation of several baselines across Multilingual performance and Cross-lingual transferability tasks.}
\label{tapp:average-all}
\end{minipage}
\end{table*}

To verify the advantages of NeuronXA in interpreting model downstream task performance and transferability, we conducted comparisons with three representation similarity-based evaluation methods. Table \ref{app-baselines} and Table \ref{app-baselines-transfer} present the correlation coefficients between alignment scores and downstream task performance, and between alignment scores and model transferability, respectively. Table \ref{tapp:average-all} shows the average correlation coefficients of all baselines with downstream task performance and model transferability. The results indicate that NeuronXA outperforms others in interpreting both multilingual capabilities and cross-lingual transferability, confirming the effectiveness and robustness of our method.

\paragraph{Centered Kernel Alignment (CKA).}
CKA \citep{kornblith2019similarity} is a similarity measure rooted in the Hilbert-Schmidt Independence Criterion (HSIC) ~\citep{gretton2005measuring}, a non-parametric method designed to assess the independence among random variables. CKA serves as a second-order similarity index, functioning by comparing the subspaces spanned by neurons, which endows it with robust power for comparing representations across different networks. Its theoretical foundation lies in identifying dominant correlation directions within distinct datasets and conducting comparisons based on these directions. Furthermore, CKA can be adjusted to a weighted version by incorporating eigenvalues, thereby giving rise to Linear CKA. By design, CKA is intended to exhibit invariance with respect to data scaling, centering, and orthogonal transformations, and it maintains its stability even under any invertible linear transformations of the data.

\paragraph{Singular Value Canonical Correlation Analysis (SVCCA).}
SVCCA is a method introduced by \citet{raghu2017svcca} for comparing learned representations in neural networks. It combines Singular Value Decomposition (SVD) and Canonical Correlation Analysis (CCA) to provide an efficient and invariant way to compare representations. The approach first applies SVD to each set of neurons to identify the most significant directions that explain the majority of the variance in the data. Then, CCA is used to find linear transformations that maximize the correlation between these subspaces from different layers or networks. SVCCA is designed to be invariant to affine transformations, making it suitable for comparisons across different architectures and training stages.

\paragraph{Averaged Neuron-Wise Correlation (ANC).}
The ANC method, introduced by ~\citet{del-fishel-2022-cross}, offers a novel approach to analyzing cross-lingual similarity in multilingual language models. It is based on the assumption that neurons in the representations of different languages are aligned one-to-one a priori. ANC calculates the correlations between pairs of neurons from different languages and then averages these correlations to generate a similarity score. Compared to other methods, ANC provides improved interpretability by enabling the identification of specific neurons that contribute the most or the least to the similarity.

\section{NeuronXA Score for Other Datasets} 
% other-benchmark
\begin{table*}[h]
\centering
% \footnotesize
% \resizebox{1.0\textwidth}{!}{
\begin{tabular}{llccccccc}
\toprule
 &  & Llama 3.1 & Llama 3 & Qwen 2.5 & Mistral 0.3 & OLMo 2 & GLM 4 & AVG \\
 &  & 8B & 8B & 14B & 7B & 7B & 9B   \\

\midrule
\multirow{9}{*}{\rotatebox{90}{m-ARC}}
\multirow{3}{*}{\rotatebox{90}{\scalebox{.65}{weighted}}}
& MEXA & \color{purple}\textbf{0.8274} & 0.8264 & 0.8140 & {\ul 0.8961} & {\ul 0.9046} & 0.8043 & 0.8455 \\
 & NASCA & 0.7197 & {\ul 0.9134} & {\ul 0.9011} & 0.8825 & 0.9046 & {\ul 0.9382} & {\ul 0.8766} \\
 & NAVCA & {\ul 0.8102} & \color{purple}\textbf{0.9364} & \color{purple}\textbf{0.9419} & \color{purple}\textbf{0.9685} & \color{purple}\textbf{0.9446} & \color{purple}\textbf{0.9447} & \color{purple}\textbf{0.9244} \\
 
\cmidrule{2-9}
\multirow{11}{*}{\rotatebox{90}{\phantom{FLORES}}}
\multirow{3}{*}{\rotatebox{90}{\scalebox{.65}{average}}}
& MEXA & \color{purple}\textbf{0.7405} & 0.7086 & 0.9036 & \color{purple}\textbf{0.9042} & 0.8783 & \color{purple}\textbf{0.9261} & \color{purple}\textbf{0.8436} \\
 & NASCA & 0.6869 & {\ul 0.8335} & \color{purple}\textbf{0.9347} & 0.7991 & {\ul 0.8795} & 0.8800 & 0.8356 \\
 & NAVCA & {\ul 0.7075} & \color{purple}\textbf{0.8464} & {\ul 0.9152} & {\ul 0.8118} & \color{purple}\textbf{0.8956} & {\ul 0.8838} & {\ul 0.8434} \\

\cmidrule{2-9}
\multirow{11}{*}{\rotatebox{90}{\phantom{FLORES}}}
\multirow{3}{*}{\rotatebox{90}{\scalebox{.65}{last}}}
& MEXA & {\ul 0.8119} & 0.8139 & 0.8172 & 0.8487 & 0.9114 & 0.8369 & 0.8400 \\
 & NASCA & 0.7200 & \color{purple}\textbf{0.9152} & {\ul 0.9121} & {\ul 0.9261} & {\ul 0.9157} & \color{purple}\textbf{0.9451} & {\ul 0.8890} \\
 & NAVCA & \color{purple}\textbf{0.8392} & {\ul 0.9134} & \color{purple}\textbf{0.9347} & \color{purple}\textbf{0.9563} & \color{purple}\textbf{0.9548} & {\ul 0.9413} & \color{purple}\textbf{0.9233} \\

\midrule
\midrule
\multirow{9}{*}{\rotatebox{90}{m-MMLU}}
\multirow{3}{*}{\rotatebox{90}{\scalebox{.65}{weighted}}}
& MEXA & 0.7644 & 0.7627 & 0.5269 & 0.7119 & {\ul 0.8272} & 0.6717 & 0.7108 \\
 & NASCA & \color{purple}\textbf{0.9155} & \color{purple}\textbf{0.9168} & \color{purple}\textbf{0.9201} & {\ul 0.7813} & \color{purple}\textbf{0.8597} & \color{purple}\textbf{0.9143} & \color{purple}\textbf{0.8846} \\
 & NAVCA & {\ul 0.9069} & {\ul 0.9086} & {\ul 0.8609} & \color{purple}\textbf{0.8706} & 0.7997 & {\ul 0.9029} & {\ul 0.8749} \\

\cmidrule{2-9}
\multirow{11}{*}{\rotatebox{90}{\phantom{FLORES}}}
\multirow{3}{*}{\rotatebox{90}{\scalebox{.65}{average}}}
 & MEXA & 0.7357 & 0.7241 & \color{purple}\textbf{0.8738} & \color{purple}\textbf{0.9295} & \color{purple}\textbf{0.8652} & \color{purple}\textbf{0.8833} & \color{purple}\textbf{0.8353} \\
 & NASCA & {\ul 0.8404} & \color{purple}\textbf{0.8398} & {\ul 0.8654} & {\ul 0.7134} & {\ul 0.8645} & {\ul 0.8508} & {\ul 0.8291} \\
 & NAVCA & \color{purple}\textbf{0.8421} & {\ul 0.8345} & 0.8601 & 0.6814 & 0.8434 & 0.8260 & 0.8146 \\

\cmidrule{2-9}
\multirow{12}{*}{\rotatebox{90}{\phantom{FLORES}}}
\multirow{3}{*}{\rotatebox{90}{\scalebox{.65}{last}}}
& MEXA & 0.7267 & 0.7246 & 0.4953 & 0.6359 & {\ul 0.8354} & 0.7223 & 0.6900 \\
 & NASCA & \color{purple}\textbf{0.9131} & \color{purple}\textbf{0.9128} & \color{purple}\textbf{0.8938} & \color{purple}\textbf{0.8608} & \color{purple}\textbf{0.8549} & \color{purple}\textbf{0.9180} & \color{purple}\textbf{0.8922} \\
 & NAVCA & {\ul 0.8742} & {\ul 0.8719} & {\ul 0.8313} & {\ul 0.8066} & 0.7684 & {\ul 0.8899} & {\ul 0.8404} \\ 
 
\midrule
\midrule
\multirow{9}{*}{\rotatebox{90}{Belebele}}
\multirow{3}{*}{\rotatebox{90}{\scalebox{.65}{weighted}}}
& MEXA & 0.6424 & 0.6553 & 0.4100 & 0.6180 & {\ul 0.9159} & 0.5104 & 0.6253 \\
 & NASCA & \color{purple}\textbf{0.8952} & \color{purple}\textbf{0.9039} & \color{purple}\textbf{0.9179} & {\ul 0.7659} & \color{purple}\textbf{0.9246} & \color{purple}\textbf{0.8360} & \color{purple}\textbf{0.8739} \\
 & NAVCA & {\ul 0.8282} & {\ul 0.8440} & {\ul 0.7941} & \color{purple}\textbf{0.8180} & 0.8120 & {\ul 0.7867} & {\ul 0.8139} \\

\cmidrule{2-9}
\multirow{11}{*}{\rotatebox{90}{\phantom{FLORES}}}
\multirow{3}{*}{\rotatebox{90}{\scalebox{.65}{average}}}
 & MEXA & 0.7977 & 0.7950 & {\ul 0.8553} & \color{purple}\textbf{0.9351} & {\ul 0.8948} & {\ul 0.8091} & \color{purple}\textbf{0.8478} \\
 & NASCA & \color{purple}\textbf{0.8571} & \color{purple}\textbf{0.8510} & \color{purple}\textbf{0.8880} & {\ul 0.7294} & \color{purple}\textbf{0.9030} & \color{purple}\textbf{0.8209} & {\ul 0.8416} \\
 & NAVCA & {\ul 0.8340} & {\ul 0.8277} & 0.8406 & 0.6927 & 0.8627 & 0.7943 & 0.8087 \\

\cmidrule{2-9}
\multirow{11}{*}{\rotatebox{90}{\phantom{FLORES}}}
\multirow{3}{*}{\rotatebox{90}{\scalebox{.65}{last}}}
& MEXA & 0.5997 & 0.6127 & 0.4006 & 0.5301 & \color{purple}\textbf{0.9147} & 0.5591 & 0.6028 \\
 & NASCA & \color{purple}\textbf{0.8903} & \color{purple}\textbf{0.9000} & \color{purple}\textbf{0.8763} & \color{purple}\textbf{0.8408} & {\ul 0.9107} & \color{purple}\textbf{0.8208} & \color{purple}\textbf{0.8732} \\
 & NAVCA & {\ul 0.7979} & {\ul 0.8101} & {\ul 0.7662} & {\ul 0.7471} & 0.7639 & {\ul 0.7689} & {\ul 0.7757} \\ 
 
 \midrule
\midrule
\multirow{9}{*}{\rotatebox{90}{BMLAMA-53}}
\multirow{3}{*}{\rotatebox{90}{\scalebox{.65}{weighted}}}
& MEXA & \color{purple}\textbf{0.7394} & {\ul 0.7402} & 0.6949 & {\ul 0.7553} & 0.7980 & 0.7283 & 0.7427 \\
 & NASCA & 0.7223 & 0.7314 & {\ul 0.8074} & 0.6982 & {\ul 0.8489} & {\ul 0.7304} & {\ul 0.7564} \\
 & NAVCA & {\ul 0.7377} & \color{purple}\textbf{0.7434} & \color{purple}\textbf{0.8359} & \color{purple}\textbf{0.8488} & \color{purple}\textbf{0.9158} & \color{purple}\textbf{0.7463} & \color{purple}\textbf{0.8047} \\

\cmidrule{2-9}
\multirow{11}{*}{\rotatebox{90}{\phantom{FLORES}}}
\multirow{3}{*}{\rotatebox{90}{\scalebox{.65}{average}}}
& MEXA & {\ul 0.7378} & 0.7238 & {\ul 0.8177} & \color{purple}\textbf{0.8390} & {\ul 0.8926} & {\ul 0.7612} & \color{purple}\textbf{0.7954} \\
 & NASCA & 0.7266 & {\ul 0.7442} & 0.8085 & {\ul 0.6698} & 0.8809 & 0.7263 & 0.7594 \\
 & NAVCA & \color{purple}\textbf{0.7450} & \color{purple}\textbf{0.7530} & \color{purple}\textbf{0.8383} & 0.6683 & \color{purple}\textbf{0.9025} & \color{purple}\textbf{0.7815} & {\ul 0.7814} \\

\cmidrule{2-9}
\multirow{11}{*}{\rotatebox{90}{\phantom{FLORES}}}
\multirow{3}{*}{\rotatebox{90}{\scalebox{.65}{last}}}
& MEXA & {\ul 0.7232} & {\ul 0.7259} & 0.6669 & 0.6794 & 0.8115 & {\ul 0.7513} & 0.7264 \\
 & NASCA & 0.7135 & 0.7106 & {\ul 0.8238} & {\ul 0.8035} & {\ul 0.8633} & 0.7349 & {\ul 0.7749} \\
 & NAVCA & \color{purple}\textbf{0.7480} & \color{purple}\textbf{0.7479} & \color{purple}\textbf{0.8333} & \color{purple}\textbf{0.8061} & \color{purple}\textbf{0.9108} & \color{purple}\textbf{0.7531} & \color{purple}\textbf{0.7999} \\
\bottomrule
\end{tabular}
% }
\caption{Pearson correlation of NeuronXA on the Tatoeba dataset across there multilingual benchmarks and one Cross-language transfer task. The values in the table represent the correlation of NeuronXA and benchmark settings. The highest average correlations for each task are highlighted in \textcolor{purple}{\textbf{bold}}, and the second highest are \underline{underlined}.}
\label{other-benchmark}
\end{table*}
We examine the model's evaluation results on other datasets, specifically using the Tatoeba dataset. Additionally, we explore the Pearson correlation coefficients between alignment scores and three multilingual benchmarks, as well as the correlation coefficients with zero-shot cross-lingual transfer performance.

As shown in Table \ref{other-benchmark}, NeuronXA achieves relatively high correlation coefficients compared to sentence embeddings, suggesting that NeuronXA is a more generalizable method that can be applied across different datasets.

It is important to note that the quality of the bilingual datasets used for NeuronXA evaluation—particularly their distribution and diversity—can influence the alignment scores. Ideally, the greater the diversity of the dataset, the more accurately NeuronXA reflects the alignment of semantic knowledge across languages. Despite the relatively lower diversity of the Tatoeba dataset, as evidenced in Table \ref{other-benchmark}, NeuronXA still achieves a reasonably high correlation coefficients, further validating the robustness of our approach.

\section{Robustness of NeuronXA}
\label{robustness}

Similar to the discussion of MEXA \cite{kargaran2024mexa}, NeuronXA scores are highly robust, with a very low probability of achieving randomly high values. Our matrix $\mu_{C(l)}$ measures the alignment scores of matrix $C(l)$, specifically the proportion of diagonal elements that attain the maximum value within their respective rows and columns. We assume the existence of an $n$-dimensional matrix $C(l)$, with $k$ elements satisfying this condition. For an $N \times N$ matrix, the probability of diagonal elements being the maximum value in both their row and column is given by $p = \frac{1}{2n-1}$.
\begin{equation}\label{randomprob}
P(X \geq \frac{k}{n}) = 1 - \sum_{i=0}^{k-1} \binom{n}{i} p^i (1 - p)^{n - i}
\end{equation}

Assuming the diagonal elements are the maximum in both their row and column, the probability that at least $k$ of the $n$ independent variables satisfy this condition can be computed using the binomial distribution formula in \ref{randomprob}. This formula suggests that, given a sufficient number of parallel sentences ($n$), the likelihood of achieving a high score by chance is very low. For example, with $n = 100$, the probability of obtaining a NeuronXA alignment score greater than 0.05 (with $k = 5$) from a random $100 \times 100$ matrix is $p(x \geq 0.05) = 0.00016$.

\end{document}